\pdfoutput=1
\documentclass[11pt]{article}

\usepackage[]{ACL2023}
\usepackage{adjustbox}
    \usepackage{times}
\usepackage{latexsym}
\usepackage{array}
\newcolumntype{L}{>{\centering\arraybackslash}m{3cm}}
\usepackage[T1]{fontenc}
\usepackage[utf8]{inputenc}
\usepackage{multirow}
\usepackage{microtype}

\usepackage{inconsolata}
\newcommand*{\affmark}[1][*]{\textsuperscript{#1}}

\title{Aesthetics of Sanskrit Poetry from the Perspective of Computational Linguistics: A Case Study Analysis on \'Sik\d{s}\={a}\d{s}\d{t}aka}

\author{Jivnesh Sandhan\affmark[1], 
Amruta Barbadikar\affmark[2], Malay Maity\affmark[2], Pavankumar Satuluri\affmark[3],\\ \textbf{Tushar Sandhan\affmark[1], Ravi M. Gupta\affmark[4], Pawan Goyal\affmark[5] and Laxmidhar Behera\affmark[1,6]}\\
\affmark[1]IIT Kanpur, \affmark[2]University of Hyderabad, \affmark[3]IIT Roorkee,  \\ \affmark[4]Utah State University, \affmark[5]IIT Kharagpur, \affmark[6]IIT Mandi \\
\texttt{jivneshsandhan@gmail.com,pawang@cse.iitkgp.ac.in}}

\begin{document}
\maketitle
\begin{abstract}
Sanskrit poetry has played a significant role in shaping the literary and cultural landscape of the Indian subcontinent for centuries. However, not much attention has been devoted to uncovering the hidden beauty of Sanskrit poetry in computational linguistics. This article explores the intersection of Sanskrit poetry and computational linguistics by proposing a roadmap of an interpretable framework to analyze and classify the qualities and characteristics of fine Sanskrit poetry. We discuss the rich tradition of Sanskrit poetry and the significance of computational linguistics in automatically identifying the characteristics of fine poetry. We also identify various computational challenges involved in this process, including subjectivity, rich language use, cultural context and lack of large labeled datasets. The proposed framework involves a human-in-the-loop approach that combines deterministic aspects delegated to machines and deep semantics left to human experts.

We provide a deep analysis of \textit{\'Sik\d{s}\={a}\d{s}\d{t}aka}, a Sanskrit poem, from the perspective of 6 prominent \textit{\textit{k\={a}vya}\'s\={a}stric} schools, to illustrate the proposed framework. Additionally, we provide compound, dependency, \textit{\textit{anvaya}} (prose order linearised form), meter, \textit{rasa} (mood), \textit{ala\.nk\={a}ra} (figure of speech), and \textit{r\={\i}ti} (writing style) annotations for \textit{\'Sik\d{s}\={a}\d{s}\d{t}aka} and a web application to illustrate the poem's analysis and annotations. Our key contributions include the proposed framework, the analysis of \textit{\'Sik\d{s}\={a}\d{s}\d{t}aka}, the annotations and the web application\footnote{Link for interactive analysis of \textit{\'Sik\d{s}\={a}\d{s}\d{t}aka}: \url{https://sanskritshala.github.io/shikshastakam/}} for future research. We aim to bridge the gap between \textit{\textit{k\={a}vya}\'s\={a}stra} and computational methods and pave the way for future research.
\end{abstract}

\section{Introduction}
Sanskrit literature has a rich and diverse tradition that has played a significant role in shaping the literary and cultural landscape of the Indian subcontinent for centuries \cite{pollock2006language,jamison2014rigveda}. With its complex grammatical rules and nuanced vocabulary, the Sanskrit language has provided a fertile ground for poets to craft intricate and evocative stanzas that capture the key elements of human experience \cite{pollock1996sanskrit}. From the ancient epics like the R\={a}m\={a}ya\d{n}a and the Mah\={a}bh\={a}rata, to the lyrical works of K\={a}lidasa and Bhart\d{r}hari etc. Sanskrit poetry has embodied the key elements of Indian thought and culture, providing inspiration and contemplation for generations of readers and scholars. However, the researchers in the computational linguistics community have not devoted much attention to uncovering the hidden beauty in Sanskrit poetry \cite{scharf2015distinctive}.
% As such, it is an important area of study for the computational linguistic research community, which should be interested in identifying the best poetry automatically. 

The computational linguistic research community could be interested in identifying fine poetry for various reasons, including the need to evaluate machine-generated poetry \cite{agarwal-kann-2020-acrostic,van-de-cruys-2020-automatic,li-etal-2020-rigid,hopkins-kiela-2017-automatically}, translating poetry \cite{yang-etal-2019-generating,ghazvininejad-etal-2018-neural,krishna-etal-2019-poetry}, to gain a deeper understanding of poetic language \cite{haider-2021-metrical,hamalainen-alnajjar-2019-lets,waltz-1975-understanding}, and to develop recommender systems \cite{10.1145/3341161.3342885,foley2019poetry} for readers interested in poetry. By analyzing large corpora of poetry \cite{haider-2021-metrical,haider-etal-2020-po,gopidi-alam-2019-computational,fairley-1969-stylistic} and identifying the most outstanding examples, researchers can improve evaluation metrics for machine-generated poetry \cite{yi-etal-2018-automatic,liu-etal-2019-rhetorically,ghazvininejad-etal-2017-hafez,goncalo-oliveira-2017-survey}, develop new models and techniques for natural language processing \cite{hu-sun-2020-generating}, and  preserve cultural heritage. Furthermore, personalized recommendations for readers can be developed based on their preferences.

Analyzing and identifying the characteristics of fine poetry automatically presents several computational challenges, including: (1) Subjectivity: The perception of what makes a poem ``good" is subjective and can vary widely among individuals and cultures \cite{sheng-uthus-2020-investigating}. Developing a computational model that can accurately capture and evaluate the aesthetic qualities of a poem is therefore challenging.
(2) Rich language use: Poetry often employs complex language use, including metaphors, similes, allusions, and wordplay, which can be difficult for computational models to understand and generate \cite{chakrabarty-etal-2021-dont}.
(3) Cultural context: Poetry is often deeply rooted in cultural and historical contexts \cite{gopidi-alam-2019-computational}, which can be challenging for automated systems to comprehend and incorporate.
(4) Lack of large labeled datasets: The development of automated systems to identify the characteristics of fine poetry relies on large, labeled datasets, which can be difficult to create due to the subjective nature of poetry and the diversity of cultural and linguistic contexts \cite{al-ghamdi-etal-2021-dependency,horvath-etal-2022-elte}.
Overall, the computational challenges of identifying the characteristics of fine poetry automatically include the development of sophisticated algorithms that can account for the subjective nature of poetry, understand complex language use, incorporate cultural context and work with limited labeled data.

In this article, we aim to address the question of whether we can build an automated and interpretable framework to analyze and classify Sanskrit poetry (\textit{k\={a}vya}) \cite{kesarwani-etal-2017-metaphor} into levels of the characteristics of fine composition. This framework has the potential to answer interesting questions such as given two k\={a}vyas, which one is more beautiful and why. It could also serve as a valuable tool for pedagogical purposes. The \textit{\textit{k\={a}vya}\'s\={a}stra} provides various perspectives for the analysis of poetry, and we select a \textit{\'Sik\d{s}\={a}\d{s}\d{t}aka} composition\footnote{Refer to the actual piece and its translation in Appendix \ref{shikastakam_composition}.} to analyze from these different perspectives. We also aim to discuss whether computational methods can aid in similar analysis and the limitations of existing state-of-the-art methods.
Our proposed framework involves a human-in-the-loop approach \cite{zhipeng-etal-2019-jiuge,ghazvininejad-etal-2017-hafez}, where deterministic aspects are delegated to machines, and deep semantics are left to human experts. We hope that with further development, this framework can eventually become fully automated, enhancing the appreciation of Sanskrit poetry's inner beauty for neophytes and Sanskrit enthusiasts alike. We believe that this work can serve as a stepping stone, bridging the gap between the tradition of \textit{\textit{k\={a}vya}\'s\={a}stra} and computational methods, and paving the way for future research.

Our key contributions are as follows:
\begin{enumerate}
    \item We propose a roadmap of human-in-loop and interpretable framework to classify Sanskrit poetry into levels of the characteristics of fine composition (\S \ref{proposed_semantic_framework}).
    \item We illustrate the proposed framework by providing a deep analysis of \textit{\'Sik\d{s}\={a}\d{s}\d{t}aka} from the perspective of 6 prominent \textit{\textit{k\={a}vya}\'s\={a}stra} schools (\S \ref{all_analysis}).
    \item We provide compound, dependency, \textit{anvaya}, meter, \textit{rasa}, ala\.{n}k{\=a}ra, \textit{r\={\i}ti} annotations for \textit{\'Sik\d{s}\={a}\d{s}\d{t}aka} (\S \ref{web_inteface}).
    \item We also provide web application\footnote{Link for interactive analysis of \textit{\'Sik\d{s}\={a}\d{s}\d{t}aka}: \url{https://sanskritshala.github.io/shikshastakam/}} to illustrate our \textit{\'Sik\d{s}\={a}\d{s}\d{t}aka} analysis and annotations (\S \ref{web_inteface}).
    \item We publicly release our codebase of framework and web-application for future research\footnote{\url{https://github.com/sanskritshala/shikshastakam}}
\end{enumerate}

\section{Background}
 \begin{table*}[t]
\centering
\begin{adjustbox}{}
\small
\begin{tabular}{|p{0.08\linewidth}|p{0.17\linewidth}|p{0.13\linewidth}|p{0.22\linewidth}|p{0.2\linewidth}|}
\hline
\textbf{School}   & \textbf{Founder}        & \textbf{Treatise}                  & \textbf{Objective}                                                                     & \textbf{English meaning}                                                                                                                                                         \\\hline
\textit{rasa}     & Bharatamun\={\i}    & \textit{n\={a}\d{t}ya\'s\={a}stra}              & \textit{na hi \textit{rasa}d\d{r}te kaścidartha\d{h} pravartate}                                         & No meaning proceeds, if a poetry does not carries \textit{rasa}.                                                                                                                      \\\hline
\textit{ala\.{n}k\={a}ra} & Bh\={a}maha        & K\={a}vy\={a}la\.{n}k\={a}ra              & \textit{rūpakādirala\d{m}kāra\d{h} tasyānyairvahudhodita\d{h} na kāntamapi nirbhū\d{s}a\d{m} vibhāti vanitāmukham } & Ala\.{n}k\={a}ras are vital for poetic beauty, just like ornaments for a charming woman's face. \\ \hline
\textit{r\={\i}ti}    & V\={a}mana         & K\={a}vy\={a}la\.{n}k\={a}ra-s\={u}trav\d{r}tti & \textit{rītirātmā kāvyasya}                                                         & Poetic style (\textit{r\={\i}ti}) is the soul of poetry.                                                                                                                                  \\\hline
\textit{dhvani}    & \={A}nandavardhana & Dhvany\={a}loka                & \textit{kāvyasyātmā \textit{dhvani}\d{h} }                                                       & \textit{dhvani} is the soul of poetry.                                                                                                                                                \\\hline
\textit{vakrokti} & Kuntaka        & \textit{vakrokti}-j\={\i}vitam         & \textit{\textit{vakrokti}\d{h} kāvyajīvitam}                                                   & \textit{vakrokti} is the vital element of the poetry.                                                                                                                                 \\\hline
\textit{aucitya} & K\d{s}emendra     & \textit{aucitya}-vic\={a}ra-carc\={a}  & \textit{\textit{aucitya}\d{m} \textit{rasa}siddhasya sthira\d{m} kāvyasya jīvitam}                                    & Propriety is the stable vital element of the poetry full of \textit{rasa}.    \\  \hline
\end{tabular}
\end{adjustbox}
\caption{Chronological overview of 6 schools of kāvyaśāstra with founder \={a}c\={a}rya, their treatise and objectives.} 
\label{table:kāvyaśāstra}
\end{table*}
\paragraph{The 6 main schools of \textit{\textit{k\={a}vya}\'s\={a}stra} (Poetics):}
\textit{\textit{k\={a}vya}\'s\={a}stra} is the traditional Indian science of poetics and literary criticism, which has played a significant role in shaping the development of literature and aesthetics in the Indian subcontinent for over two thousand years. The term ``\textit{k\={a}vya}'' refers to poetry or literature, while ``\textit{\'S\={a}stra}'' means science or knowledge, and \textit{\textit{k\={a}vya}\textit{\'s\={a}stra}} is thus the systematic study of the nature, forms, and principles of poetry and literature.
The roots of \textit{k{\=a}vya\'s\={a}stra} can be traced back to ancient India, where it developed alongside other branches of learning such as philosophy, grammar, and rhetoric. Over time, \textit{k\={a}vya}\'{s}\={a}stra evolved into a complex and sophisticated system of poetics, encompassing a wide range of concepts and techniques for analyzing and appreciating poetry, such as \textit{rasa} (mood of poetry), \textit{ala\.{n}k\={a}ra} (the use of rhetorical and figurative devices), \textit{dhvani} (superiority of suggestive meaning), \textit{vakrokti} (oblique), \textit{aucitya} (appropriateness) and \textit{r\={\i}ti} (style of writing). 
% Table \ref{table:kāvyaśāstra} gives a brief overview of these 6 schools in chronological order with founder \={a}c\={a}rya, their treatise and objectives.

\paragraph{\textit{chanda\'s\'s\={a}stra}} 
 is the traditional Indian science of meter and versification in poetry, dating back to the Vedic period. It involves the systematic study of the principles, forms, and structures of meter and versification in poetry, including the use of various poetic devices and the study of various types of meters and poetic forms. While closely related to \textit{\textit{k\={a}vya}\'s\={a}stra}, \textit{chanda\'s\'s\={a}stra} is a separate branch of traditional Indian knowledge and is not considered one of the 6 main schools of \textit{\textit{k\={a}vya}\'s\={a}stra}. Key concepts and techniques include the classification of meters, rhyme and alliteration, and the principles of accentuation and stress. \textit{chanda\'s\'s\={a}stra} has played a significant role in the development of Indian poetics and literary theory, and continues to be a vital part of Indian cultural heritage.
By mastering the principles of \textit{chanda\'s\'s\={a}stra}, poets and writers can create verses that are both aesthetically pleasing and technically precise, resulting in some of the most beautiful and evocative poetry in the world.

\paragraph{Efforts put in Computational Linguistics:}
In the field of computational poetry, researchers have tackled various problems using algorithms and computational methods, such as generating new poetry \cite{agarwal-kann-2020-acrostic,van-de-cruys-2020-automatic,li-etal-2020-rigid,hopkins-kiela-2017-automatically,krishna-etal-2020-graph}, translating poetry \cite{yang-etal-2019-generating,ghazvininejad-etal-2018-neural} analyzing emotional tone, analyzing rhyme and meter patterns \cite{kao-jurafsky-2012-computational,greene-etal-2010-automatic,fang-etal-2009-adapting}, classifying poetry \cite{baumann-etal-2018-analysis,kesarwani-etal-2017-metaphor}, and recommending poetry \cite{10.1145/3341161.3342885,foley2019poetry}.
However, existing work does not focus much on analyzing the characteristics of fine poetry, which is crucial for improving the generation, translation, and recommendation applications. This gap motivates our work, which proposes an interpretable framework to classify Sanskrit poetry into different levels of  composition using computational linguistics.
We demonstrate the proposed framework by conducting a deep analysis of \textit{\'Sik\d{s}\={a}\d{s}\d{t}aka}, a Sanskrit poem, from the perspective of 6 well-known \textit{\textit{k\={a}vya}\'s\={a}stric} schools. The key contributions of our article include the proposed framework, the analysis of \textit{\'Sik\d{s}\={a}\d{s}\d{t}aka}, the annotations, the web application, and the publicly released codebase for future research.

\section{Basics of schools, Computational Aspect, Open Issues and Analysis}
\label{all_analysis}
This section presents the basics of the 7 schools of \textit{\textit{k\={a}vya}\'s\={a}stra} and \textit{chanda\'s\'s\={a}stra}. Each subsection provides a comprehensive overview of the respective school's fundamentals, the feasibility of developing a computational system, open issues and future directions.
We provide reasoning for choosing the \textit{\'Sik\d{s}\={a}\d{s}\d{t}aka} and demonstrate its analysis based on the corresponding school. 
\paragraph{Why \textit{\'Sik\d{s}\={a}\d{s}\d{t}aka}:}
To facilitate a more thorough and nuanced analysis of \textit{k\={a}vya}, we set a few criteria to guide our selection of the most appropriate composition for our study. These criteria are as follows: (1) the composition should be sufficiently small to enable comprehensive scrutiny from diverse \textit{\textit{k\={a}vya}\'s\={a}stra} perspectives; (2) the poet who has authored the composition ought not to possess a significant reputation in the \textit{\textit{k\={a}vya}\'s\={a}stra} domain to obviate the potential for biases that may arise from establishing the composition as \textit{uttama} \textit{k\={a}vya}; (3) the composition should have made a substantive contribution to the traditions from a non-literary perspective to explore the role of \textit{\textit{k\={a}vya}\'s\={a}stra} aspects in its accomplishment.

The \textit{\'Sik\d{s}\={a}\d{s}\d{t}aka} composition is well-suited for our analysis as it satisfies all three criteria that we have established. First, it meets the requirement of being compact enough to allow for in-depth examination from various \textit{\textit{k\={a}vya}\'s\={a}stra} perspectives as it consists of only eight stanzas.\footnote{We encourage readers to go through this composition: \url{https://sanskritshala.github.io/shikshastakam/shloka.html} where word-to-word meanings and translations are given.} Additionally, not all \textit{a\d{s}\d{t}akas} can be considered \textit{uttama} \textit{k\={a}vya} (the fine poetry), which further supports its appropriateness as a subject of study.
Second, the author of \textit{\'Sik\d{s}\={a}\d{s}\d{t}aka}, Caitanya Mah\={a}prabhu, is not primarily known as a poet or literary figure. Thus, he meets the second criterion of not being a well-established poet in the \textit{\textit{k\={a}vya}\'s\={a}stra} domain, which helps to minimize the potential biases that could arise from claiming the composition as \textit{uttama} \textit{k\={a}vya}. 
Finally, \'Sik\d{s}\={a}\d{s}\d{t}aka has made significant contributions to the traditions of Gau\d{d}\={\i}ya Vai\d{s}\d{n}avism from a non-literary perspective, which satisfies the third criterion. 
This composition has significantly influenced the lifestyle of the religious community adhering to the principles of Gaudiya Vaishnavism, serving as the fundamental backbone of their philosophy for the past 500 years. The enduring prevalence and practice of these teachings within the community underscore the profound impact of this composition on their way of life.
% The composition emphasizes the importance of devotional service to God, humility, compassion, and spiritual love, and has become a crucial component of the practices and beliefs of the \textit{gau\d{d}\={\i}}ya Vai\d{s}\d{n}avism tradition. Thus, the \textit{\'Sik\d{s}\={a}\d{s}\d{t}aka}'s rich cultural and religious significance allows for an investigation of whether \textit{\textit{k\={a}vya}\'s\={a}stra} aspects contributed to its achievement.
In summary, the \textit{\'Sik\d{s}\={a}\d{s}\d{t}aka} composition satisfies all three criteria.

\subsection{Metrical analysis}
\paragraph{Basics:} The bulk of Sanskrit literature consists of poetic works that conform to the conventions of Sanskrit prosody, or \textit{chanda\'s\'s\={a}stra}, which involves the study of Sanskrit meters, or \textit{chandas}. The primary objective of utilizing \textit{chandas} is to infuse rhythm into the text to facilitate memorization, while also aiding in the preservation of accuracy to some degree \cite{deo_2007,Melnad}. Pi\.{n}gala is considered to be the father of \textit{chanda\'s\'s\={a}stra}. Prominent scholars and their work in this field include Pi\.{n}gala (\textit{Chanda\'s\'s\={u}tra}), Bharata Mun\={\i} (\textit{\textit{n\={a}\d{t}ya\'s\={a}stra}}), Pi\.{n}gala Bha\d{t}\d{t}a (\textit{Chandomanjar\={\i}}), V\d{r}ttan\={a}tha (\textit{V\d{r}ttaratn\={a}kara}), Jagann\={a}tha Pa\d{n}\d{d}ita (\textit{rasaga\.{n}g\={a}dhara}), etc. Although approximately 1,398 meters are described in V\d{r}ttaratn\={a}kara \cite{Rajagopalan2018AUT}, not all are widely used.
% Previous works on the theory of Chanda\d{h}\'s\={a}stra and automated tools have been conducted by \newcite{deo_2007,Melnad,neill-2023-skrutable,Rajagopalan2018AUT,terdalkar-bhattacharya-2023-chandojnanam}. 

\paragraph{Computational aspect:} In \textit{chanda\'s\={a}stra}, the classification of each syllable (\textit{ak\d{s}ara}) in Sanskrit as either \textit{laghu} (short) or \textit{guru} (long) enables the representation of every line of Sanskrit text as a binary sequence comprised of \textit{laghu} and \textit{guru} markers. The identification of specific patterns within these sequences leads to the recognition of various types of \textit{chandas}, thus enabling the deterministic identification of a meter. In recent years, commendable efforts have been made by the computational community to develop user-friendly toolkits for meter identification \cite{neill-2023-skrutable,Rajagopalan2018AUT,terdalkar-bhattacharya-2023-chandojnanam}.

\paragraph{\textit{\'Sik\d{s}\={a}\d{s}\d{t}aka} analysis:} The \textit{a\d{s}\d{t}aka}, which comprises a series of 8 stanzas, is typically composed in a single meter. In the case of the \textit{\'Sik\d{s}\={a}\d{s}\d{t}aka}, there are 5 meters employed, namely, \textit{\'sārdūlavikrī\d{d}ita, 
vasantatilaka, anu\d{s}\d{t}up, viyogin\={\i}} and \textit{upaj\={a}ti} (\textit{indrava\.m\'sa} and \textit{va\d{m}\'sastha}).
Further details on how to identify meters can be found in the relevant literature \cite{neill-2023-skrutable,Rajagopalan2018AUT,terdalkar-bhattacharya-2023-chandojnanam}. The identification of meters has aided in the identification of typographical errors (\textit{tanuja} $\rightarrow$ \textit{tanūja},  \textit{dhūli}$\rightarrow$ \textit{dhūlī}, \textit{gadagada}$\rightarrow$ \textit{gadgada}) in some of the existing manuscripts of the \textit{\'Sik\d{s}\={a}\d{s}\d{t}aka}. One may wonder about the use of 5 different meters in the \textit{\'Sik\d{s}\={a}\d{s}\d{t}aka}. Some scholars, or \textit{\={a}c\={a}ryas}, have argued that the stanzas were not composed in a single sitting. Rather, they were collected in Rupa Goswami's (disciple of the author of \textit{\'Sik\d{s}\={a}\d{s}\d{t}aka}) Pady\={a}val\={\i}, which is an anthology of compositions by expert poets in the \textit{gau\d{d}\={\i}}ya Vai\d{s}\d{n}ava tradition. K\d{r}\d{s}\d{n}adas Kaviraj Goswami later compiled and arranged them in a meaningful order in his composition, the Caitanya Carit\={a}m\d{r}ta. We posit that the use of multiple meters serves as an evidence supporting this claim.

\paragraph{Open issues and future direction:} It would be interesting to explore the correlation between the intrinsic mood of the meter and the mood of the poetry. It would also be worthwhile to investigate the relationship between poets and the meters they have employed. This could involve examining patterns in their favorite choices and attempting to identify the poet based on the composition itself.  Additionally, there is potential for exploring whether the occurrence of 
\textit{rasado\d{s}a} (obstruction in the enjoyment of mellow) can be predicted through automatic identification of the composition's \textit{chandas}.  These investigations could contribute to a deeper understanding of the relationship between meters, poetry, and emotion in Sanskrit literature.

\subsection{\textit{ala\.{n}k\={a}ra} school}
\paragraph{Basics:}  This school focuses on the use of figurative language or literary devices to enhance the beauty and aesthetic appeal of poetry. It includes the study of metaphors, similes, personification, hyperbole, and other literary devices. The most important exponent of this school is Bh\={a}maha, who wrote the \textit{k\={a}vy\={a}la\.{n}k\={a}ra}, one of the earliest treatises on poetic embellishments. Other notable figures associated with this school include Da\d{n}\d{d}in, Udbha\d{t}a, Rudra\d{t}a and so on. There are two broad categories of \textit{ala\.{n}k\={a}ras}: (1) \textit{\textit{\'{s}abd\={a}la\.{n}k\={a}ra}}: refers to the figure of sound that relies on the pleasing sound or choice of words, which loses its effect when the words are substituted with others of similar meaning.  (2) \textit{\textit{arth\={a}la\.{n}k\={a}ra}}: is the figure of speech that is based on the meaning of words.

\paragraph{Computational aspect:} The process of identifying \textit{\'{s}abd\={a}la\.{n}k\={a}ra} is considered deterministic and involves verifying the occurrence of specific patterns of syllables or words. In contrast, the identification of \textit{arth\={a}la\.{n}k\={a}ra} presents a significant semantic challenge, even for experienced annotators. The feasibility and difficulty of developing supervised systems for this purpose can be better understood through empirical investigations.
There is no system currently available for automated \textit{ala\.{n}k\={a}ra} analysis for Sanskrit. To develop a supervised data-driven system, several important questions must be addressed. For instance, it is crucial to determine the amount of data that needs to be annotated, as well as the appropriate methods for marking \textit{ala\.{n}k\={a}ra} annotations, such as whether they should be marked at the level of a complete \'{s}loka and a phrase.
To address these concerns, it is necessary to develop a standardized scheme for \textit{ala\.{n}k\={a}ra} annotation. This would enable researchers to collect and analyze annotated data consistently and systematically, which would facilitate the development of accurate and reliable automated systems for \textit{ala\.{n}k\={a}ra} analysis.

\paragraph{\textit{\'Sik\d{s}\={a}\d{s}\d{t}aka} analysis:} In our analysis of the \textit{\'Sik\d{s}\={a}\d{s}\d{t}aka}, we have employed the \textit{ala\.{n}k\={a}ra} categorization outlined by Mamma\d{t}a in his influential work ``K\={a}vyaprak\={a}\'{s}a". This text provides a comprehensive overview of 67 \textit{ala\.{n}k\={a}ras} or literary ornaments that can be utilized to embellish and enhance the expression of poetry. According to the \textit{ala\.{n}k\={a}ra} school, \textit{ala\.{n}k\={a}ra} is considered the soul of poetry, and without its incorporation, a poetic composition lacks vitality. In other words, \textit{k\={a}vya} without \textit{ala\.{n}k\={a}ra} is likened to a lifeless entity, deemed by some poets of this school to be comparable to a widow.\footnote{\textit{ala\.{n}kārarahitā vidhavaiva sarasvatī|} 2.334, Agnipur\={a}\d{n}a\\ A poetry which is also a form of goddess Sarasva\={\i}, is not decorated by \textit{ala\.nk\=aras}, looks like a widow women, who is not wearing any ornaments.} 
Furthermore, it is worth noting that certain poets from this tradition place greater emphasis on \textit{arth\={a}la\.{n}k\={a}ra}, or figurative language, over \textit{\'{s}abd\={a}la\.{n}k\={a}ra}, which deals primarily with sound patterns and word repetition. The \textit{\'Sik\d{s}\={a}\d{s}\d{t}aka}, a devotional composition in Sanskrit, exhibits the mood of separation (\textit{vipralambha-\'{s}\d{r}\.{n}gāra\d{h}}) as its primary \textit{rasa} or mood. The author portrays himself as a devotee, and his beloved is Lord Govinda. The appropriate use of ala\.{n}k\={a}ras (figures of speech) in this composition serves to express the \textit{rasa} without hindrance. The author employs \textit{\'{s}abd\={a}la\.{n}k\={a}ras} (\textit{anupr\={a}sa}\footnote{\textit{var\d{n}as\=amyamanupr\=asa\d{h}}| 9.79, K\={a}vyaprak\={a}\'{s}a \\ The similarity in the letters used in the poetry is `Anupr\=asa'. It is similar to the alliteration in the English language.}) and \textit{arth\={a}la\.{n}k\={a}ras} (\textit{r\={u}paka}\footnote{\textit{tadr\=upakambhedo ya\d{h} upam\=anopameyayo\d{h}|} 10.93, K\={a}vyaprak\={a}\'{s}a\\ R\={u}paka is a figure of speech, where the upameya, which is being compared and the upam\=ana with which is compare are explained as if both are not different from each other. It is similar to the metaphor.}, \textit{upam\={a}}\footnote{\textit{s\=adharmyamupam\=a bhede|}1.87, K\={a}vyaprak\={a}\'{s}a \\ When the upameya and the upam\=ana are not compared based on the similarity possessed by the both is called as Upam\=a. It is similar to the simile.}, \textit{vyatireka}\footnote{\textit{upam\=an\=ad yadanyasya vyatireka\d{h} sa\d{h} eva sa\d{h}}, 10.105, K\={a}vyaprak\={a}\'{s}a \\ If the upameya is said to be having a higher degree of quality than the upam\=ana, then it is called as vyatireka \textit{ala\.nk\=ara}.}, \textit{tulyayogitā}\footnote{\textit{niyat\=an\=a\.m sak\d{r}ddharma\d{h} s\=a puna\d{h} tulyayogit\=a|} 10.104, K\={a}vyaprak\={a}\'{s}a\\ A single mention of a property belonging to various other things is Tulyayogit\=a} and \textit{viśe\d{s}okti}\footnote{\textit{viśe\d{s}oktirakha\d{n}\d{d}e\d{s}u k\=ara\d{n}e\d{s}u phalāvaca\d{h}|}
10.108, K\={a}vyaprak\={a}\'{s}a \\ The condition in which various reasons are stated but the result is not seen is called vi\'se\d{s}okti. There are two types of this figure of speech. 1) - when the reason is specified for the unseen effect it is \textit{uktanimitta-vi\'se\d{s}okti} otherwise it is \textit{anuktanimitta-vi\'se\d{s}okti.}} ) in \textit{\'Sik\d{s}\={a}\d{s}\d{t}aka}. The use of \textit{r\={u}paka} (metaphor)  and \textit{upam\={a}} is frequent, with the former appearing 6 times and the latter 4 times. The metaphors are employed to equate two things directly, while similes use ``like" or ``as" to make comparisons.
 
In the first stanza, the author uses a series of 5 metaphors to describe the victory of \textit{\'{S}rik\d{r}\d{s}\d{n}a} \textit{Sa{\.n}k\={\i}rtana}. The use of these metaphors creates a garland-like effect, which serves as an offering to acknowledge the victory of \'{S}r\={\i}k\d{r}\d{s}\d{n}a \textit{Sa{\.n}k\={\i}rtana}. Additionally, the use of \textit{anty\={a}nupr\=asa \'Sabdala\.nk\=ara} (a figure of speech where every word ends with a similar suffix) creates a rhythmic effect. %Although, antyanup\textit{rasa} is not mentioned in kavyaprakasha 
In the second stanza, the author expresses his sorrow of not having an attachment to such a glorious \textit{Sa{\.n}k\={\i}rtana}, where the Lord invests all his potencies without keeping any rules and regulations. Although there is every reason to get attached to such \textit{Sa{\.n}k\={\i}rtana}, the author cannot get attached, which is expressed through \textit{Uktanimitta vi\.se\d{s}yokti \textit{ala\.nk\=ara}.} This \textit{ala\.{n}k\={a}ra} is used to denote that all the causes are present for action to happen, yet the action does not occur.
In the third stanza, the author talks about the prerequisites of developing attachment for the \textit{Sa\.nk\={\i}rtana}: humility and tolerance. The author employs \textit{vyatireka} \textit{ala\.nk\=ara} to embellish this stanza by comparing humility with a grass and tolerance with a tree. In this figure of speech, the subject of illustration (upam\=ana) has higher qualities than the subject of comparison (upameya). Thus, the author says that one should be lower than a grass and more tolerant than a tree.
In the fourth stanza, multiple things (wealth, followers, beautiful women and flowery language of Vedas) are coupled with a single property ``\textit{na k\=amaye}'' (do not desire). This device is called \textit{tulyayogitā} \textit{ala\.{n}k\={a}ra}.
In the fifth stanza, the author beautifully couples metaphor and simile to express his eternal servitude to  Govinda. The author compares the way water drops fall from a lotus into the mud, similarly, he has fallen from Govinda's lotus feet to the ocean of nescience (material existence) as a dust-like servant. The author requests Govinda to place him back at his lotus feet.
In the sixth stanza, again \textit{tulyayogitā} \textit{ala\.nk\=ara} is employed to connect different symptoms mentioned to a single property.
In the seventh stanza, the author expresses his intense separation from Govinda by deploying three similes (\textit{upam\=a} \textit{ala\.nk\=ara}). The author uses these similes to express his love in the mood of separation. He compares a moment to a great millennium, his eyes to the rainy season, and the entire world to void. This garland of similes (\textit{m\={a}lopam\=a} \textit{ala\.nk\=ara}) serves as an offering to express the author's love for Govinda.
The final stanza employs the \textit{anuktanimitta-viśe\d{s}okti} device, wherein a condition is expressed but the consequent effect is absent, and no explanation is offered for the lack of effect. This constitutes the \textit{anukatanimitta-viśe\d{s}okti} \textit{ala\.nk\=ara}. The author enumerates several potential causes for being despised, yet the anticipated outcome of animosity does not manifest. Despite hatred conditions, the author maintains devotion to the lord. The rationale for the absence of hatred is not mentioned here.
In conclusion, the \textit{\'Sik\d{s}\={a}\d{s}\d{t}aka} employs a range of ala\.{n}k\={a}ras to express its mood (\textit{rasa}) without hindrance. The appropriate use of metaphors, similes, and other figures of speech serves to create a garland-like effect that acknowledges Govinda's glory and expresses the author's devotion and love.

\paragraph{Open issues and future direction:} Moving forward, several directions for research on \textit{ala\.nk\=aras} can be explored. One crucial area is the formulation of the problem of identifying \textit{ala\.nk\=aras}. Since \textit{ala\.nk\=aras} can be assigned to complete or partial stanzas, or sequences of syllables or pairs of words, determining the appropriate approach for identifying them is non-trivial. Different formulations, such as sentence classification, sequence labeling, or structured prediction, can be employed.
Another interesting direction for research is to investigate which ala\.{n}k\={a}ras are more beautiful and how we can compare two compositions based solely on their use of \textit{ala\.nk\=aras}. Defining a basis for evaluating beauty in poetry is a challenging problem in itself. However, understanding which ala\.{n}k\={a}ras contribute more to the aesthetic appeal of a poem and how to measure this appeal can be valuable for poets and scholars alike.
Finally, exploring the correlation of \textit{ala\.{n}k\={a}ra} school with other schools of \textit{\textit{k\={a}vya}\'s\={a}stra}, such as \textit{rasa} and \textit{dhvani}, is a promising area for future research. Understanding how these different aspects of poetry interact can provide insight into the complex mechanisms of poetic expression and perception. Overall, these future directions offer exciting opportunities to deepen our understanding of \textit{ala\.nk\=aras} in poetry.

\subsection{\textit{rasa} school}
\paragraph{Basics:}  The \textit{rasa} school of Indian aesthetics places significant emphasis on the emotional impact of poetry on the reader or audience. Bharata mun\={\i}, the author of the  \textit{n\={a}\d{t}ya\'s\={a}stra}, is considered the most influential thinker associated with this school. His treatise on Indian performing arts includes a detailed discussion of \textit{rasa} theory, which posits that \textit{rasa} is the soul of poetry. According to Bharata, \textit{rasa} is the ultimate emotional pleasure that can be derived from a work of art. 
\textit{rasa} is the heightened emotional response to the text. Bharata also classified the various emotions depicted in performing arts into eight different \textit{rasa}s, or flavors, which are \textit{\'s\d{r}\.ng\=ara} (romance), \textit{h\=asya} (comedy), \textit{karu\d{n}a} (piteous), \textit{raudra} (anger), \textit{v\={\i}ra} (heroism), \textit{bhay\=anaka} (fear), \textit{b\={\i}bhatsa} (disgust), and \textit{adbhuta} (wonder). The ninth emotion, \textit{\'s\=anta} (peace) is also enlisted by the other scholars.\footnote{\textit{na yatra du\d{h}kha\d{m} na sukha\d{m} na dve\d{s}o nāpi matsara\d{h}| sama\d{h} sarve\d{s}u bhūte\d{s}u sa śānta\d{h} prathito \textit{rasa}\d{h} ||}, \textit{n\={a}\d{t}ya\'s\={a}stra} .\\Where there is absence of sorrow, joy, hatred, and jealousy, with existing sense of equality for everyone is the \textit{`śānta} \textit{rasa}'.} The \textit{\'s\=anta} \textit{rasa} is not compatible for dramas but other kinds of poetry can be composed in this \textit{rasa}. 
The concept of \textit{rasa} theory remains an integral part of Indian culture and has greatly influenced the performing arts in India and beyond.

In Bharata's  \textit{n\={a}\d{t}ya\'s\={a}stra}, a formula\footnote{\textit{vibhāvānubhāvavyabhicāri\-sa\d{m}yogād\textit{rasa}ni\d{s}patti\d{h}|} ,\textit{n\={a}\d{t}ya\'s\={a}stra} \\The \textit{rasa} is the combined effect of \textit{vibhāvā}, \textit{anubhāva}, vyabhicāribh\=ava} for the arousal of \textit{rasa} is presented, wherein the combination of \textit{vibh\=ava}, \textit{anubh\=ava}, and \textit{vyabhic\=ar\={i} \textit{bh\=ava}} \textit{bh\=ava} is necessary to evoke the experience of \textit{rasa}. The term `\textit{vibh\=ava}' refers to the stimulants of emotions, while `\textit{anubh\=ava}' represents the physical responses that accompany emotional reactions, and `\textit{vyabhic\=ar\={\i}} \textit{bh\=ava}' denotes the transitory emotions. The basic emotions or bh\=avas of the reader or spectator are activated by the vibh\=avas, while the anubh\=avas and \textit{vyabhic\=ar\={\i}} bh\=avas serve to indicate the emotional response experienced. This formula suggests that the experience of \textit{rasa} is not solely dependent on a single aspect of the literary text or performance, but rather requires the combination of multiple elements to evoke an emotional response.

\paragraph{Computational aspect:} Computational identification of \textit{rasa} in poetry  is an interesting and challenging research area. 
One of the main challenges in identifying \textit{rasa} computationally is that the interpretation of \textit{rasa} is subjective. Different readers or critics may have different opinions on the dominant emotion evoked by a particular work of poetry.
Another challenge is the complexity of the \textit{vibh\=ava}, \textit{anubh\=ava}, and \textit{vyabhic\=ar\={\i}}-\textit{bh\=ava} that contribute to the formation of \textit{rasa}. These elements are often implicit or subtle in the text, and their identification requires a deep understanding of the cultural and literary context in which the text was produced.
To overcome these challenges, possible computational approaches that can be taken include using machine learning algorithms to identify patterns in the text that are associated with particular emotions, and incorporating contextual information such as the author, genre, and historical period. In addition, the use of computational linguistics techniques, such as sentiment analysis and topic modeling, may also be helpful in identifying the dominant emotions expressed in a text.
However, it is important to note that while computational systems may be able to identify the dominant \textit{rasa} of a work of poetry, they may not be able to fully capture the richness and complexity of the \textit{vibh\=ava}, \textit{anubh\=ava}, and \textit{vyabhic\=ar\={\i}}-\textit{bh\=ava} that contribute to the formation of \textit{rasa}. The identification of these elements requires a deep understanding of the cultural and literary context in which the text was produced, and is likely to remain the purview of human literary critics and scholars.
In conclusion, while computational approaches may be able to provide some insight into the identification of \textit{rasa} in poetry, the complexity and subjectivity of \textit{rasa} theory poses significant challenges. Further research in this area is needed to develop computational models that can better capture the cultural and literary context of poetic works and the nuances of the \textit{vibh\=ava}, \textit{anubh\=ava}, and \textit{vyabhic\=ar\={\i}} \textit{bh\=ava} that contribute to the formation of \textit{rasa}.

\paragraph{\textit{\'Sik\d{s}\={a}\d{s}\d{t}aka} analysis:} A fundamental aspect of evoking \textit{rasa} involves combining three elements, namely \textit{vibh\=ava}, \textit{anubh\=ava}, and \textit{vyabhic\=ar\={\i}} \textit{bh\=ava}. The \textit{vibh\=avas} activate the basic emotions or \textit{bh\=avas} of the reader or spectator, while the \textit{anubh\=avas} and \textit{vyabhic\=ar\={\i}} \textit{bh\=avas} indicate the emotional response experienced. In the composition under discussion, the author assumes the mood of a devotee and reciprocates with his beloved Lord K\d{r}\d{s}\d{n}a. The author employs several words to indicate that the \textit{vibh\=ava} is \textit{k\d{r}\d{s}\d{n}a}, including \textit{śrī-k\d{r}\d{s}\d{n}a-sa\.{n}kīrtanam} (congregational chanting of K\d{r}\d{s}\d{n}a's holy name), \textit{nāmnām} (K\d{r}\d{s}\d{n}a's holy name), \textit{hari\d{h}} (another name of K\d{r}\d{s}\d{n}a), \textit{nanda-tanuja} (K\d{r}\d{s}\d{n}a, the son of Nanda Maharaj), \textit{tava nāma-graha}\d{n}e (chanting K\d{r}\d{s}\d{n}a's holy name), Govinda (another name of K\d{r}\d{s}\d{n}a), and \textit{mat-prā\d{n}a-nātha\d{h}} (the master of my life).
The \textit{anubh\=ava} of the author is evident in the seventh stanza, where he expresses intense feelings of separation. He considers even a moment to be a great millennium, and tears flow from his eyes like torrents of rain. He perceives the entire world as void. The author employs several words to indicate the \textit{vipralambha s\d{r}\.ng\=ara}, such as \textit{nānurāga\d{h}} (no attachment to K\d{r}\d{s}\d{n}a), \textit{a-darśanāna} (K\d{r}\d{s}\d{n}a not being visible), \textit{govinda-virahe\d{n}a} (separation from Govinda), and \textit{mat-prā\d{n}a-nātha\d{h}} (the master of my life). In conclusion, it is evident from our analysis of \textit{vibh\=ava} and \textit{anubh\=ava} in this composition that the \textit{rasa} evoked is \textit{vipralambha s\d{r}\.ng\=ara}, which is characterized by intense feelings of separation and longing. The author effectively employs \textit{vibh\=avas}, such as the names and descriptions of K\d{r}\d{s}\d{n}a, to activate the basic emotions of the reader or spectator, while his \textit{anubh\=avas}, such as his intense feelings of separation and tears, serve to indicate the emotional response experienced. The composition is a powerful example of how the combination of \textit{vibh\=ava} and \textit{anubh\=ava} can be used to evoke \textit{rasa} in Indian classical literature.

\paragraph{Open issues and future direction:} The future direction of computational linguistics in identifying \textit{rasa} in poetry requires a comprehensive approach that covers all the crucial dimensions of the problem. 
Data annotation is one crucial aspect of the future direction of computational linguistics in identifying \textit{rasa} in poetry. Data annotation involves the manual tagging of texts with information such as the dominant \textit{rasa}, \textit{vibh\=ava}, \textit{anubh\=ava}, and \textit{vyabhic\=ar\={\i}}\textit{bh\=ava}. 
Possible approaches that can be taken include using supervised machine learning algorithms to identify patterns in the text that are associated with particular emotions. This involves training a model on a labeled dataset of texts annotated with the dominant \textit{rasa} and associated \textit{vibh\=ava}, \textit{anubh\=ava}, and \textit{vyabhic\=ar\={\i}}\textit{bh\=ava}. 
Another approach is to use unsupervised machine learning algorithms such as clustering and topic modeling to identify the dominant emotions expressed in the text. This approach requires minimal human annotation.
In addition, incorporating contextual information such as the author, genre, and historical period can also be useful in identifying the dominant \textit{rasa} in a text.

\subsection{\textit{r\={\i}ti} school}
\paragraph{Basics:} In this study, the focus is on the concept of \textit{r\={\i}ti}, also referred to as \textit{m\=arga} (way), in poetry. This aspect of poetry has been heavily emphasized by the scholar V\=amana, who considers \textit{r\={\i}ti} as the soul of poetry. According to V\=amana, there are three \textit{r\={\i}tis}, namely \textit{vaidarbh\={\i}}, \textit{gau\d{d}\={\i}}, and \textit{p\=a\~nc\=al\={\i}}, which were named after the geographical areas from where poets constructed in these styles. The deterministic elements of \textit{r\={\i}ti} include the letters, length and quantity of compounds, and the complexity or simplicity of the composition.
The \textit{vaidarbh\={\i}} style is considered the most delightful style of poetry, consisting of all ten \textit{gu\d{n}as}\footnote{\textit{samagragu\d{n}opet\=a \textit{vaidarbh\={\i}}}| 2.11, K\={a}vy\={a}la\.{n}k\={a}ra-s\={u}trav\d{r}tti\\ The composition style having all the qualities is \textit{vaidarbh\={\i}}} explained by V\=amana. The construction of \textit{vaidarbh\={\i}} contains a shorter compounds.
\footnote{We need to perform empirical analysis of existing annotated poetry to get threshold of length of short compounds.} 
The \textit{gau\d{d}\={\i}} style, dominated by \textit{oja} and \textit{k\=anti }\textit{gu\d{n}as},
\footnote{\textit{\textit{oja}\d{h}k\=antimat\={\i} \textit{gau\d{d}\={\i}}y\=a}| 2.12, K\={a}vy\={a}la\.{n}k\={a}ra-s\={u}trav\d{r}tti\\\textit{oja} and \textit{k\=anti }are the qualities (\textit{gu\d{n}as}) that exist in the construction with long compounds and complex arrangement of letters. When these qualities are explicit in the writing style, the \textit{r\={\i}ti} is called as \textit{gau\d{d}\={\i}}.} and lacks \textit{sukum\=arat\=a} and \textit{m\=adhurya} \textit{gu\d{n}as}. \textit{sukum\=arat\=a} and \textit{m\=adhurya} are the qualities which occur in a construction having less compounds and soft letter combinations are used. But, here in the \textit{gau\d{d}\={\i}}, the composition possesses long compounds and too many joint consonants. 
\textit{p\=a\~nc\=al\={\i}} style has a simple composition having no or less compounds and simple construction.

V\=amana also describes ten qualities of sound and ten qualities of meaning, including \textit{\'sle\d{s}a} (close association), \textit{\textit{pras\={a}da}} (simplicity), \textit{samat\=a} (uniformity of diction), \textit{m\=adhurya} (sweetness or pleasantness), \textit{sukumarat\=a} (softness), \textit{arthavyakti} (explicitness), \textit{ud\=arata} (raciness), \textit{oja} (floridity), \textit{k\=anti }(brilliancy), and \textit{sam\=
adhi} (symmetry).\footnote{\textit{\textit{oja}\d{h} p\textit{rasa}daśle\d{s}asamatāsamādhimādhurya\-saukumāryodāratārthavyaktikāntayo bandhagu\d{n}ā\d{h}||} 3.4, K\={a}vy\={a}la\.{n}k\={a}ra-s\={u}trav\d{r}tti\\\'Sle\d{s}a, \textit{pras\={a}da}, Samat\=a, \textit{m\=adhurya}, Sukumarat\=a, Arthavyakti, Ud\=arata, \textit{oja}, K\=anti, and Sam\=
adhi are the 10 qualities of diction.} These 10 qualities of poetry are compressed into 3 qualities suggested by letters, compounds, and diction.
\textit{m\=adhurya} (pleasantness), which leads to the melting of the mind, is suggestive of the \textit{spar\'sa} consonants (k, kh, g, gh, c, ch, j, jh, t, th, d, dh, p, ph, b, bh) with the exception of those in the `\d{t}a' class (\d{t}, \d{t}h, \d{d}, \d{d}h), combined with nasel consonants (\.{n}, ñ, n, m), the consonants `r' and  `\d{n}' when followed by a short vowel, and when there are no or medium length compounds. \textit{oja}, the cause of lustrous expansion of mind, is suggestive of complex diction consisting of combinations of first consonants of the class (k, c, \d{t}, t, p) with the second consonants (kh, ch, \d{t}h, th, ph) and the third (g, j, \d{d}, d, b)  with the fourth one gh, jh, \d{d}h, dh, bh), any consonant with `r' in any order, any consonant with itself, the `\d{t}a' class without `\d{n}', `ś' and `\d{s}', and consisting of long compounds. \textit{pras\={a}da} is present in all the \textit{rasa}s and all kinds of poetries and \textit{r\={\i}tis} equally. In \textit{pras\={a}da} gu\d{n}a, words can be comprehended right after the hearing of it.
In conclusion, \textit{r\={\i}ti} plays an essential role in the composition of poetry, as emphasized by V\=amana. \textit{vaidarbh\={\i}}, \textit{gau\d{d}\={\i}}, and \textit{p\=a\~nc\=al\={\i}} are the three \textit{r\={\i}tis}, each having its own unique features and characteristics. Additionally, the ten qualities of sound and ten qualities of meaning and their compression into three \textit{gu\d{n}as} suggested by letters, compounds, and diction are important elements of \textit{r\={\i}ti}. The proper deployment of these elements can significantly enhance the impact of poetry on the reader or listener.

\paragraph{Computational aspect:} The identification of the \textit{r\={\i}ti} of a composition can be considered a deterministic process. \textit{\textit{k\={a}vya}\'s\={a}stra} provides various clues that enable the classification of a \textit{r\={\i}ti} category. For instance, if a composition comprises of soft syllables, no joint consonants, and a dearth of compounds or short compounds, it is considered as \textit{vaidarbh\={\i}} \textit{r\={\i}ti}. Conversely, if a composition contains more joint consonants, long compounds, the usage of fricatives, and specific combinations of joint syllables, it is considered as \textit{gau\d{d}\={\i}} \textit{r\={\i}ti}. The remaining compositions, which do not fall into these two categories, are classified as \textit{p\=a\~nc\=al\={\i}} \textit{r\={\i}ti}. To the best of our knowledge,  currently no computational system exists that can determine the \textit{r\={\i}ti} of a composition.

\paragraph{\textit{\'Sik\d{s}\={a}\d{s}\d{t}aka} analysis:} The \textit{\'Sik\d{s}\={a}\d{s}\d{t}aka} is an example of a composition that embodies the \textit{m\=adhurya} gu\d{n}a and \textit{vaidarbh\={\i}} \textit{r\={\i}ti}. This is apparent through the simple composition, usage of soft consonants, limited occurrence of long compounds, and the restricted usage of joint consonants. Moreover, the \textit{rasa} of the \textit{k\={a}vya} is \textit{vipralambha \'s\d{r}\.ng\=ara}, which is aptly brought out through the use of \textit{vaidarbh\={\i}} \textit{r\={\i}ti} and \textit{m\=adhurya} \textit{gu\d{n}a}. The utilization of any other \textit{r\={\i}ti} would have led to a \textit{rasa} \textit{do\d{s}a}, highlighting the significance of choosing the appropriate \textit{r\={\i}ti} to convey the intended \textit{rasa}.

\paragraph{Open issues and future direction:} A natural direction for future research would be to consolidate the provided clues and develop a rule-based computational system for the identification of \textit{r\={\i}ti} in various compositions by different poets. This would enable automated analysis and assessment of \textit{r\={\i}ti} in \textit{k\={a}vya}. An empirical investigation could be carried out to explore the degree to which \textit{r\={\i}ti} is effective in identifying the \textit{rasa} of a \textit{k\={a}vya}. Additionally, further research could explore the correlation of \textit{r\={\i}ti} school with other \textit{\textit{k\={a}vya}\'s\={a}stra} schools. V\=amana, the founder of the \textit{r\={\i}ti} school, maintains that different \textit{r\={\i}ti} can provide unique experiences for readers. Therefore, it would be of interest to conduct a cognitive analysis of brain signals of readers after exposing them to compositions of different \textit{r\={\i}tis}. This would provide valuable insights into the effect of \textit{r\={\i}ti} on the reader's cognitive processes and emotional responses to \textit{k\={a}vya}.

\subsection{\textit{dhvani} school}
\paragraph{Basics:} The school of \textit{dhvani}, a prominent school of Indian poetics, is associated with \=Anandavardhana, who authored the seminal work Dhvany\=aloka, which expounds on the theory of suggestion in poetry. Other significant figures in this school include Abhinavagupta, Mamma\d{t}a, etc. . According to the \textit{dhvani} school, poetry's key elements lies in suggestion or \textit{dhvani}, which alludes to the verbal, hidden, and profound aspects of poetry. This school emphasizes the power of suggestion and implication in poetry and argues that a poem's true meaning is not merely in its literal content, but in the emotional response it elicits in the reader through suggestion and implication.
The meaning is distinguished in three categories : (1) \textit{v\=acya} or literal meaning, (2) \textit{lak\d{s}a\d{n}a} or indirect meaning, which is preferred when the literal meaning is inadequate and contradictory, and (3)\textit{ vya\.ngya} or poetic/metaphysical meaning, which is a deeper meaning that does not conflict with the literal meaning and evokes wonder by reaching out to the soul of poetry. When the \textit{ vya\.ngy\=artha} is felt more delightful than the other meanings, the \textit{ vya\.ngy\=artha} is called as \textit{dhvani}.\footnote{\textit{yatrārtha\d{h} śabdo vā tamarthamupasarjanīk\d{r}tasvārthau |
vya\.{n}kta\d{h} kāvyaviśe\d{s}a\d{h} sa dhvaniriti sūribhi\d{h} kathita\d{h}||} 1.13, Dhvany\=aloka\\(When the word and its meaning in a poetry, bestow to obtain the suggestive sense, which arouses the pleasure, is called as the \textit{dhvani} by the learned men.)} Based on the degree of \textit{dhvani}, a poem is classified into three categories that demonstrate high, medium, or low conformity with the standards of fine Sanskrit poetry, as articulated by experts in \textit{\textit{k\={a}vya}\'s\={a}stra}: (1) \textit{uttama} \textit{k\={a}vya} or high composition, (2) \textit{madhyama} \textit{k\={a}vya} or medium composition, and (3) \textit{adhama} \textit{k\={a}vya} or low composition.
In \textit{uttama} \textit{k\={a}vya}, \textit{dhvani} is present, evoking wonder in the reader. In \textit{madhyama} \textit{k\={a}vya}, \textit{ vya\.ngy\=artha} is present, but it does not evoke wonder compared to \textit{v\=acy\=artha}, while in \textit{adhama} \textit{k\={a}vya}, \textit{dhvani} is absent. 
Furthermore, \textit{dhvani} is divided into three broad categories: (1) \textit{vastu-dhvani}, which implies some rare fact or idea, (2) \textit{\textit{ala\.nk\=ara}-dhvani}, which suggests a figure of speech, and (3) \textit{rasa}-\textit{dhvani}, which evokes a feeling or mood in the reader.

\paragraph{Computational aspect:} Identifying \textit{dhvani} in \textit{k\={a}vya} computationally is a complex and challenging task due to the subjective nature of literary interpretation and the intricacy of the language. \textit{dhvani}, which encompasses various layers of meaning, including literal, indirect, and metaphoric, presents a challenge to computational identification due to  the absence of annotated data. Poetic devices such as simile, metaphor, and alliteration add to the complexity of creating a unified set of features for the identification of \textit{dhvani}. Furthermore, identifying \textit{dhvani} requires additional information, such as the context, mood of the composer, his biography, and his other compositions. The cultural and historical context further complicates the identification of \textit{dhvani}, which often requires scholars' commentaries to decode the layers of meanings. Even humans may not be able to identify \textit{dhvani} without these inputs, indicating the level of expertise required for such tasks. At present, to the best of our knowledge, no system is available to provide \textit{dhvani} analysis for Sanskrit poetry. 

\paragraph{\textit{\'Sik\d{s}\={a}\d{s}\d{t}aka} analysis:} The \textit{\'Sik\d{s}\={a}\d{s}\d{t}aka} is a set of 8 stanzas containing sublime instructions delivered without any specific audience. In contrast to the Bhagavad-g\={\i}t\=a, which is directed towards a particular audience, the \textit{\'Sik\d{s}\={a}\d{s}\d{t}aka} is meant for inner contemplation and self-instruction. 
We can draw three possible dhvanis from these 8 stanzas. \textit{The first \textit{dhvani} deals with the \textit{\textit{sambandha}, \textit{abhidheya}}, and \textit{prayojana} for a devotee.} \textit{sambandha} refers to the relation of the subject with the \textit{abhidheya} of devotion, \textit{abhidheya} refers to the means of devotion which is being stated, and \textit{prayojana} refers to the fruit of devotion. The first five stanzas of the \textit{\'Sik\d{s}\={a}\d{s}\d{t}aka} discuss the \textit{sambandha} aspect, while the entire set of stanzas serves as the \textit{abhidheya}. The last three stanzas focus on \textit{prayojana}, which involves absorption in the remembrance of K\d{r}\d{s}\d{n}a and the intense emotions that this experience brings.

\textit{The second \textit{dhvani} presents the higher stages of devotional service.} The first five stanzas describe \textit{s\=adhana bhakti}, or the devotional service that is practiced through discipline and rules. The last 3 stanzas discuss \textit{prema}, the ultimate perfection of devotion.
\textit{The third \textit{dhvani} is the biography of the author.} In the initial stanza of the \textit{\'Sik\d{s}\={a}\d{s}\d{t}aka}, the author's proficiency in Sanskrit grammar and poetry is conveyed by the phrase ``\textit{vidyā-vadhū-jīvanam.}" The Caitanya Caritam\d{r}ta (CC) further provides a detailed description of the author's logical expertise, wherein he is seen to skillfully refute and reestablish his own arguments, as well as his prowess in teaching devotion to K\d{r}\d{s}\d{n}a through grammar. Additionally, CC narrates an incident where the author defeated the great poetry scholar Ke\'sava K\=a\'smir\={\i}. The phrase ``\textit{para\.m vijayate śrī-k\d{r}\d{s}\d{n}a-sa\.{n}kīrtanam}" illustrates the author's victorious preaching of the message of \textit{sa\.{n}kīrtanam}, or congregational chanting of the holy names of K\d{r}\d{s}\d{n}a. CC even goes on to mention how animals like tigers, deers and elephants were inspired to participate in \textit{sa\.{n}kīrtanam}, resulting in a blissful embrace between a tiger and a deer.
The second stanza of the \textit{\'Sik\d{s}\={a}\d{s}\d{t}aka} reveals the author's egalitarian nature, as he considered everyone to be eligible to participate in \textit{sa\.{n}kīrtanam} without any discrimination. The third stanza introduces the author's \textit{b\={\i}ja-mantra}, which urges one to develop a taste for \textit{sa\.{n}kīrtanam} and show mercy towards all living entities. The author's adherence to the principle of humility is evident in his desire for devotees to follow the same. The fourth stanza recounts the author's renunciation of wealth, followers, and even his beautiful wife at the age of 24 to embrace the life of a \textit{sany\=as\={\i}}, or a renunciant. The last three stanzas of the \textit{\'Sik\d{s}\={a}\d{s}\d{t}aka} reflect the author's mood of separation in the last 24 years of his life. The CC provides a vivid description of the same. In summary, the author's teachings in the \textit{\'Sik\d{s}\={a}\d{s}\d{t}aka} are a reflection of his own life experiences and practices.

\paragraph{Open issues and future direction:} In future research, it is crucial to address the challenges of identifying suggestive meaning in poetry by formulating a well-defined problem statement. One potential direction is to explore the use of machine learning techniques to identify \textit{dhvani} in \textit{k\={a}vya}. This could involve developing annotation schemes that capture different levels of meaning, including literal, indirect, and metaphoric, which can be used to train machine learning models.
Evaluation of performance is another important direction for future research. Metrics could be developed to measure the accuracy of the system's predictions of \textit{dhvani}. These metrics could be based on human evaluations or derived automatically by comparing the system's output to expert annotations.
One addional aspect that could be explored in future research is the use of multi-lingual and cross-lingual approaches for identifying \textit{dhvani}. Many works of Sanskrit poetry have been translated into other languages, and it would be interesting to investigate whether the same \textit{dhvani} can be identified across languages, and whether insights from one language can be used to improve the identification of \textit{dhvani} in another language.
Lastly, researchers could explore ways of integrating multiple sources of information to improve the accuracy of \textit{dhvani} identification. For example, incorporating biographical information about the composer, their other works, and the cultural and historical context of the composition could help to disambiguate multiple possible meanings and identify the intended \textit{dhvani}.

\subsection{\textit{vakrokti} school}
\paragraph{Basics:} Kuntaka's \textit{vakrokti} school is a prominent school of Indian poetics that emphasizes the use of oblique expressions in poetry. Kuntaka's seminal work, Vakroktij\={i}vita, lays out the principles of this school and classifies the levels of expression of \textit{vakrokti} into six categories. 
The first category is phonetic figurativeness, which involves the skillful use of syllables to embellish the sound of the poem.
This is closely related to \textit{r\={\i}tis} and \textit{anupr\=asa} (alliteration) \textit{ala\.{n}k\={a}ra}, which are types of \textit{\'sabdala\.nk\=ara}. 
The second category is Lexical figurativeness, which involves the use of oblique expressions at the level of the root word without suffix. The third category is the obliqueness in the suffixes. The fourth category is involves the figurativeness, which involves the use of oblique expressions at the level of the sentence. 
In this category, one has to rely on a complete sentence to understand the oblique meaning. Different \textit{arthala\.nk\=aras} are categoriesed under this category. The fifth category is sectional figurativeness, which involves the twist in the chapter-wise arrangement of the poetry. The sixth category is compositional figurativeness, which involves understanding the oblique meaning by relying on the complete composition. 
Kuntaka considers \textit{vakrokti} to be the soul of poetry, and argues that it is the only embellishment possible to the word and its meaning. The use of \textit{vakrokti} is essential to create effective poetry, and mastery of the six levels of expression is fundamental to the creation of effective poetry in the \textit{vakrokti} school.

\paragraph{Computational aspect:} Upon analysis of Kuntaka's \textit{vakrokti} school, it becomes apparent that his theory of oblique expression incorporates various elements of the previously established schools of \textit{r\={\i}ti}, \textit{rasa}, \textit{dhvani}, and \textit{ala\.{n}k\={a}ra}. Despite this, Kuntaka's contribution to the field lies in his systematic classification of 6 levels of oblique expression. However, the computational challenges posed by the deep semantics of this theory are similar to those discussed in the other schools. Therefore, it is difficult to develop a module capable of accurately identifying \textit{vakrokti} in a composition. To date, no computational system exists for computing \textit{vakrokti}, as it remains a complex task.

\paragraph{\textit{\'Sik\d{s}\={a}\d{s}\d{t}aka} analysis:} The present analysis examines the use of oblique expressions in \textit{\'Sik\d{s}\={a}\d{s}\d{t}aka}. The study identifies all 6 categories of oblique expressions deployed in the poem. The first category, phonetic figurativeness, concerns the skillful use of syllables to enhance the poetic sound, and we find that \textit{\'Sik\d{s}\={a}\d{s}\d{t}aka} employs various types of \textit{anupr\=asa} \textit{ala\.nk\=aras}.
The second category, word-root level figurativeness, relates to the use of oblique expressions at the level of individual words, and the poem utilizes \textit{r\=upak} \textit{ala\.{n}k\={a}ra} for this purpose. The third category, sentential figurativeness, focuses on oblique expressions at the level of sentences, and employs several \textit{arthala\.{n}k\={a}ras}. The fourth category, Contextual Figurativeness, relies on the poem's context and utilizes a series of \textit{r\=upaka ala\.{n}k\={a}ras} to create a garland-like effect. The fifth category, compositional figurativeness, involves understanding the oblique meaning by examining the poem's different section. \textit{\'Sik\d{s}\={a}\d{s}\d{t}aka} uses this category to discuss stages of \textit{bhakti} and the concept of \textit{sambandha}, \textit{abhidheya}, and \textit{prayojana} in an oblique manner. The final category, refers to the use of complete composition to convey oblique. We find that the author talks about his biography in the entire composition in an oblique manner. Overall, the study reveals that \textit{\'Sik\d{s}\={a}\d{s}\d{t}aka} employs all six categories of oblique expressions, highlighting the author's mastery of poetic devices and his ability to convey complex spiritual ideas in a highly nuanced and layered manner.

\paragraph{Open issues and future direction:} As the field of NLP advances, there is growing interest in exploring the use of oblique expressions in computational models. In particular, the study of oblique expressions can contribute to the development of natural language generation, sentiment analysis, and machine translation systems.
One potential area of research is the identification and classification of oblique expressions in large corpora of texts. Machine learning algorithms can be trained to recognize patterns of oblique expression usage and distinguish between different types of oblique expressions, such as the six categories identified in the present analysis of \textit{\'Sik\d{s}\={a}\d{s}\d{t}aka}. This would require the creation of annotated datasets that can be used for training and testing purposes.
Another area of research is the development of computational models that can generate oblique expressions in natural language. Such models could be trained to generate oblique expressions based on the context of the text and the intended meaning. This would require a deep understanding of the different types of oblique expressions and their functions, as well as the ability to generate language that is both grammatically correct and semantically appropriate.
Despite the potential benefits of using oblique expressions in computational models, there are also several challenges that need to be addressed. One of the main challenges is the ambiguity of oblique expressions, which can make it difficult for computational models to accurately interpret their meaning. Another challenge is the variability of oblique expression usage across different languages and cultural contexts, which requires a more nuanced and context-specific approach.
% Overall, the study of oblique expressions from a computational perspective holds great promise for advancing our understanding of language and communication, but also presents several challenges that must be addressed through interdisciplinary research and collaboration between linguists, computer scientists, and other experts.

\subsection{\textit{aucitya} school}
\paragraph{Basics:} K\d{s}emendra, a renowned poet of the 11th century, introduced the concept of \textit{aucitya} in his seminal work \textit{aucitya}-vic\=ara-carc\=a. \textit{aucitya}, meaning appropriateness, suggests that the elements of poetry, such as \textit{rasa}, \textit{ala\.{n}k\={a}ra}, and \textit{r\={\i}ti}, should be used in an appropriate manner.\footnote{\textit{ucitasthānavinyāsādala\.{n}k\d{r}tirala\.{n}k\d{r}ti\d{h} | \-aucityādacyutā nitya\d{m} bhavantyeva gu\d{n}ā gu\d{n}ā\d{h}||\\(The figures are decorative when they do not overpass the appropriateness in the poetry. The qualities can be assumed as qualities in real sense when they follow the validity of appropriteness.)} 1.6, \textit{aucitya}-vic\=ara-carc\=a} \textit{aucitya} is considered the soul of a poem, and appropriateness is regarded as the secret of beautiful composition. This theory has been widely accepted by scholars without any argument and is often referred to as the `Theory of Coordination' since it regulates all aspects of  \textit{n\={a}\d{t}ya\'s\={a}stra}.
\textit{aucitya} is a tool that aids writers in wielding their poem according to their will and effectively delivering their ideas. K\d{s}emendra categorized \textit{aucitya} into several classes, namely \textit{pada} (word form), \textit{v\=akya} (sentence), \textit{prabandhan\=artha} (meaning of the whole composition), \textit{gu\d{n}a}(qualities), \textit{ala\.{n}k\={a}ra} (poetic figures), \textit{rasa} (sentiments), etc.
In conclusion, K\d{s}emendra's concept of \textit{aucitya} has had a significant impact on the field of poetry. The idea of appropriateness and the use of poetry elements in a balanced manner are essential components of successful composition. The categorization of \textit{aucitya} into multiple classes provides a comprehensive framework for analyzing and understanding the nature of poetry. Therefore, \textit{aucitya} is an invaluable tool for writers, enabling them to deliver their ideas efficiently.

\paragraph{Computational aspect:} According to the available literature, no computational system currently exists that provides an analysis of \textit{aucitya}. In order to develop a module for \textit{aucitya} analysis, we propose considering the mutual compatibility of \textit{r\={\i}ti}, \textit{rasa}, and \textit{\textit{ala\.nk\=ara}} in the initial implementation. For instance, \textit{vaidarbh\={\i}} is considered a compatible \textit{r\={\i}ti} when deploying the s\d{r}\.ng\=ara \textit{rasa}, while \textit{gau\d{d}\={\i}} is appropriate for the \textit{v\={\i}ra} \textit{rasa}. Additionally, certain \textit{ala\.nk\=aras} are suitable for specific \textit{rasa}s. At a later stage, we may annotate corpora and build data-driven metrics to automatically measure the \textit{aucitya} of a composition. However, to accomplish this, it is essential to establish standardized data annotation policies in advance. One major challenge in evaluating poetic appropriateness across various languages, cultures, and time periods is the lack of standardized rules or models. Furthermore, the subjective nature of poetic appropriateness makes it challenging to reach a consensus among scholars and experts on the definition of \textit{aucitya}.

\paragraph{\textit{\'Sik\d{s}\={a}\d{s}\d{t}aka} analysis:} Our analysis suggests that the \textit{\'Sik\d{s}\={a}\d{s}\d{t}aka} is an exemplary composition that demonstrates \textit{aucitya}. The composition employs the \textit{vipralambha s\d{r}\.ng\=ara} \textit{rasa}, and the \textit{vaidarbh\={\i}} \textit{r\={\i}ti} with the \textit{m\=adhurya} gu\d{n}a, which are perfectly compatible. Additionally, the use of \textit{ala\.{n}k\={a}ras} is deployed in an appropriate manner, without overwhelming the expression of the \textit{rasa}. The composition's \textit{rasa} remains consistent and does not change abruptly. Instead, it gradually intensifies throughout the composition, reaching its pinnacle in the last stanza. This uniformity in the \textit{rasa}'s expression contributes to the effectiveness of the composition in evoking the desired emotional response.

\paragraph{Open issues and future direction:} Moving forward, the development of a computational system for \textit{aucitya} analysis presents an exciting area of research. In terms of future directions, several questions can be explored. Firstly, the effectiveness of the proposed mutual compatibility approach between \textit{r\={\i}ti}, \textit{rasa}, and \textit{\textit{ala\.nk\=ara}} needs to be evaluated by analyzing the performance of the initial module. Furthermore, additional features could be incorporated to improve the accuracy of \textit{aucitya} analysis, such as word choice, meter, and rhyme patterns.
Moreover, there is a need to establish standardized guidelines for the annotation of corpora in different languages and cultural contexts. This would facilitate the creation of a large, diverse dataset that can be used to train the \textit{aucitya} analysis system. Additionally, research could focus on the development of data-driven metrics that automatically evaluate poetic appropriateness.
Another potential direction for research is to investigate the interplay between language, culture, and poetic appropriateness. This involves analyzing the extent to which poetic conventions vary across different languages and cultures and how these variations affect the evaluation of \textit{aucitya}. The impact of historical and temporal changes on poetic conventions can also be explored to understand the evolution of poetic conventions over time.
Finally, collaboration between experts in literature and computer science can help establish a consensus on what constitutes \textit{aucitya}. This could lead to the development of a standardized set of rules and models for evaluating poetic appropriateness, which could be applied across different languages, cultures, and time.

\section{The proposed framework}
\label{proposed_semantic_framework}
\begin{figure*}[!tbh]
    \centering
    \includegraphics[width=0.7\textwidth]{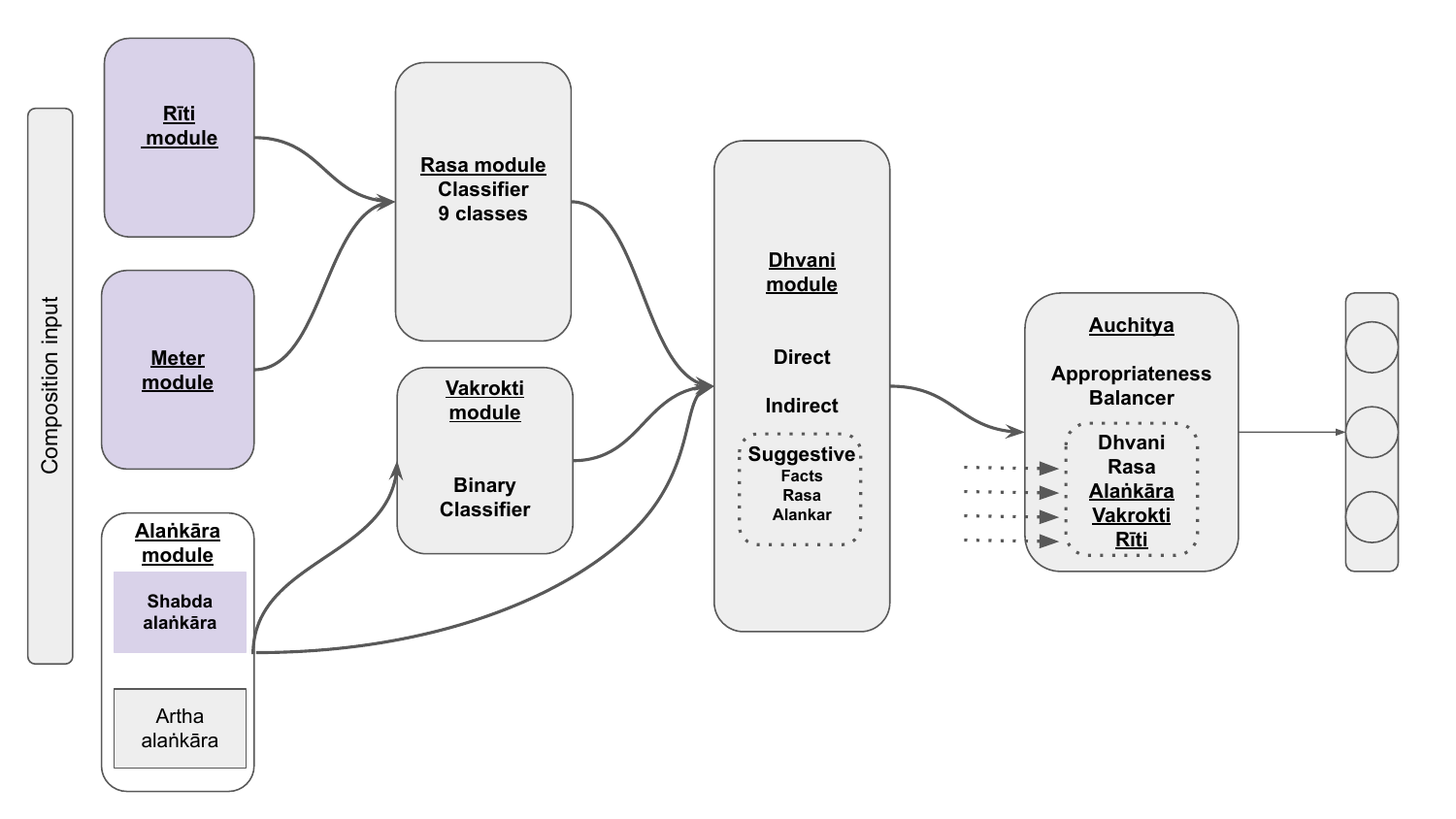}
    \caption{The proposed framework for analysis and classification of poetry using a hierarchical ensembling architecture. It consists of 7 modules, with violet modules being deterministic and the rest being supervised modules, currently supported by human experts. All modules are trained independently. The outputs and features of previous modules are helpful for the next phase modules, and the \textit{aucitya} module learns the weight of each module using all their outputs except the meter module. The composition is finally classified into three classes that demonstrate high, medium, or low conformity with the standards of fine Sanskrit poetry, as articulated by experts in \textit{\textit{k\={a}vya}\'s\={a}stra}.  We illustrate the proposed framework by showing analysis on \textit{\'Sik\d{s}\={a}\d{s}\d{t}aka} composition.}
    %convert to IAST 
    \label{figure:kāvya_framework}
\end{figure*}
This section presents the description of a proposed framework for the analysis and classification of poetry. The framework is designed to categorize compositions into 3 classes that demonstrate high, medium, or low conformity with the standards of fine Sanskrit poetry, as articulated by experts in \textit{\textit{k\={a}vya}\'s\={a}stra}. Figure \ref{figure:kāvya_framework} depicts the proposed framework, which employs a hierarchical ensembling architecture consisting of 7 modules. The violet modules are deterministic, while the remaining modules are learning-based and currently supported by human experts. Each module can be trained independently, and the outputs and features of the previous modules are utilized to enhance the performance of the next phase modules. The \textit{aucitya} module is responsible for learning the weight of each module by employing all their outputs, except the meter module. We illustrate the proposed framework by showing analysis on \textit{\'Sik\d{s}\={a}\d{s}\d{t}aka} composition.

The first layer of the framework consists of 3 modules namely, \textit{r\={\i}ti}, Meter and \textit{ala\.{n}k\={a}ra} module.
The \textit{r\={\i}ti} module is the first module in the proposed framework, and it is responsible for identifying the \textit{r\={\i}ti} of a composition. This is deterministic, and the classification is enabled by various clues provided by the \textit{\textit{k\={a}vya}\'s\={a}stra}. The next module, the meter identification module, is also deterministic, and commendable efforts have been made by the SCL community in recent years to develop user-friendly toolkits for the identification of meter \cite{neill-2023-skrutable,Rajagopalan2018AUT,terdalkar-bhattacharya-2023-chandojnanam}.
In the \textit{ala\.{n}k\={a}ra} module, the identification of \textit{\'{s}abdala\.nk\=ara} is considered deterministic, and it involves verifying the occurrence of specific patterns of syllables or words. In contrast, the identification of \textit{arthala\.nk\=ara} presents a significant semantic challenge, even for experienced annotators. The module relies on a supervised paradigm of binary classification to identify whether any \textit{arthala\.nk\=ara} is present or not.
The \textit{rasa} and \textit{vakrokti} (oblique) modules are the next two modules in the framework. For the \textit{rasa} module, a classification problem with 9 classes of \textit{rasa}s is framed. These \textit{rasa}s are \textit{\'s\d{r}\.ng\=ara} (love and beauty), \textit{h\=asya} (comedy), \textit{karu\d{n}a} (piteous), \textit{raudra} (anger), \textit{v\={\i}ra} (heroism), \textit{bhay\=anaka} (fear), \textit{b\={\i}bhatsa} (disgust), \textit{adbhuta} (wonder), and \textit{\'s\=anta} (peace). It is argued that \textit{r\={\i}ti} and meter may be helpful and can serve as auxiliary information to identify \textit{rasa}; therefore, they are used as features in the \textit{rasa} module. The \textit{vakrokti} module is framed as a binary classification problem to identify the presence of oblique. Here, the \textit{ala\.{n}k\={a}ra} module has a strong correlation in contributing to oblique identification.
The next layer of the module is the \textit{dhvani} module, which is also a classification problem with 3 classes: direct meaning, indirect meaning, and suggestive meaning. Suggestive meaning has a strong correlation with \textit{rasa}, \textit{ala\.{n}k\={a}ra}, and \textit{vakrokti}. Therefore, these features are plugged into the \textit{dhvani} module. Finally, the \textit{aucitya} module learns the weight of each module using a supervised paradigm. These weighted features of all modules are used for the final classification of the composition into three classes that demonstrate high, medium, or low conformity with the standards of fine Sanskrit poetry, as articulated by experts in \textit{\textit{k\={a}vya}\'s\={a}stra}.
In the absence of annotated data, human experts are relied upon for the modules that are supervised modules. In the future, it is hoped that all the modules can be automated, and researchers can consider training these modules in an end-to-end fashion.
Overall, the proposed framework presents a systematic approach to analyze and classify poetry into different categories based on various features. The framework has the potential to assist poets, scholars, and critics in the evaluation of compositions and provide insights into the nuances of poetry.
\begin{figure*}[!tbh]
    \centering
    \includegraphics[width=0.9\textwidth]{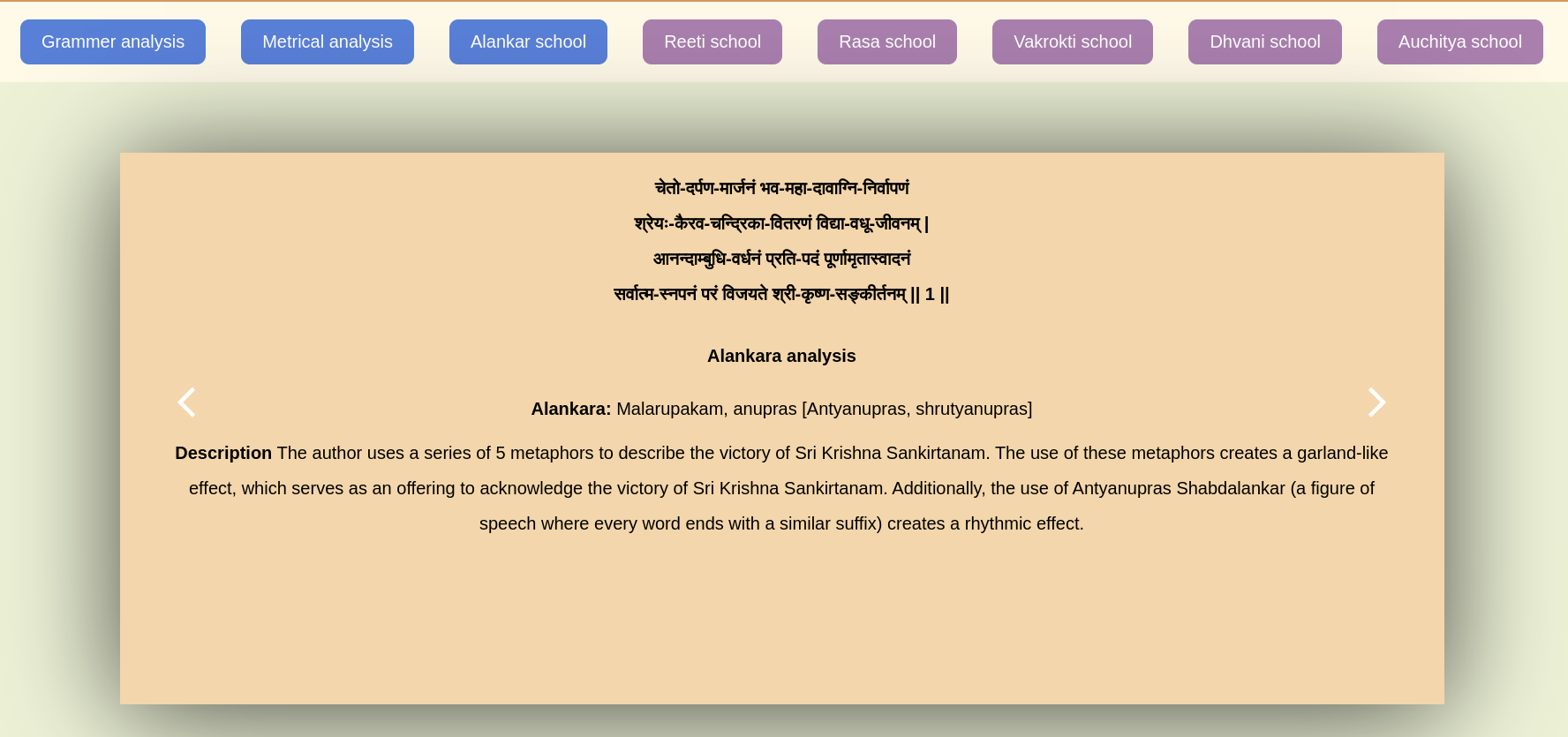}
    \caption{Web-based interactive modules to illustrate annotations and poetry analysis. Here, we showcase the verse-level analysis for \textit{ala\.nk\=ara} module.}
    \label{figure:website}
\end{figure*}
\paragraph{Is \textit{\'Sik\d{s}\={a}\d{s}\d{t}aka} an \textit{uttama} \textit{k\={a}vya}?} The criterion for a composition to be qualified as an \textit{uttama} \textit{k\={a}vya} is the presence of \textit{dhvani}, which evokes wonder in the reader.\footnote{\textit{idamuttamamatiśayini vya\.{n}gace vācyād dhvanirbudhai\d{h} kathita\d{h}|}1.4, K\={a}vyaprak\={a}\'{s}a\\ The highest kind of poetry is the one in which the suggestive meaning dominates the other meanings. In this case, the suggestive meaning is called \textit{dhvani} by scholars.} \textit{dhvani} is the soul of a \textit{k\={a}vya} and is of three types: \textit{vastu}-\textit{dhvani}, \textit{ala\.nk\=ara}-\textit{dhvani}, and \textit{rasa}-\textit{dhvani}. Vastu-\textit{dhvani} signifies rare facts or ideas; \textit{ala\.nk\=ara}-\textit{dhvani} indicates figures of speech, and \textit{rasa}-\textit{dhvani} evokes feelings or moods in the reader.
In the context of \textit{vastu}-\textit{dhvani}, \textit{\'Sik\d{s}\={a}\d{s}\d{t}aka} explores the \textit{sambandha}, \textit{abhidheya}, and \textit{prayojana} for a devotee, gradations of devotional service, and the biography of the author. The composition provides an insightful glimpse into the key elements of \textit{bhakti} and its various stages, making it a source of rare knowledge for the reader.
In terms of \textit{ala\.nk\=ara}-\textit{dhvani}, \textit{\'Sik\d{s}\={a}\d{s}\d{t}aka} employs a range of figures of speech such as metaphors, similes, and other literary devices to express its mood (\textit{rasa}) without hindrance. The appropriate use of these devices creates a garland-like effect that acknowledges Govinda's glory and expresses the author's devotion and love. The composition's poetic beauty is further enhanced by its rhythm and musicality, making it a delight for readers.
Moreover, \textit{\'Sik\d{s}\={a}\d{s}\d{t}aka} is a powerful example of how the combination of \textit{vibh\=ava} and \textit{anubh\=ava} can be used to evoke vipralambha s\d{r}\.ng\=ara \textit{rasa}, making it an exemplar of \textit{rasa}-\textit{dhvani}. The composition conveys the feeling of separation from the divine, which is central to the Vai\d{s}\d{n}ava tradition of \textit{bhakti}, and generates a deep emotional response in the reader.
In conclusion, the evidence presented above establishes that \textit{\'Sik\d{s}\={a}\d{s}\d{t}aka} is indeed an \textit{uttama} \textit{k\={a}vya}. Its exploration of rare knowledge, its \textit{ala\.{n}k\={a}ra} beauty, and its evocation of deep emotions through \textit{rasa}-\textit{dhvani} which renders greater amusement than the other meanings make it a masterpiece of Sanskrit literature.

\section{Web Interface}
\label{web_inteface}
This work employs compound tagging, morphological tagging, \textit{K\=araka} dependency tagging, and \textit{\textit{anvaya}} to annotate the \textit{\'Sik\d{s}\={a}\d{s}\d{t}aka}. The primary objective of this annotation exercise is to facilitate a literal understanding of the composition. By providing grammatical information, this annotation serves as a baseline to evaluate the \textit{dhvani}, or poetic suggestion, of the composition. The stanzas were annotated by 4 annotators, each an expert in one of the aforementioned tasks. In cases where there was a discrepancy in the annotation process, the annotators discussed and resolved the issue. All annotators hold a minimum academic qualification of a Master in Arts in Sanskrit. The annotations are publicly available via an interactive web-based platform as well as standard text annotation format.

\paragraph{Web-based application}
Figure \ref{figure:website} shows our web application that offers an interactive module to illustrate annotations and poetry analysis.
The motivation to build a web-based application to illustrate poetry analysis \cite{delmonte-2015-visualizing,delmonte-prati-2014-sparsar} is that it can help to make poetry analysis more accessible, engaging, and effective, providing students with a rich learning experience that encourages deeper understanding and appreciation of this art form. This template can serve as a proxy to illustrate how an automated system should produce the poetry analysis. 
There are 7 dimensions to our analysis which can be classified into two broad categories, namely, stanza-level (metrical and \textit{ala\.nk\=ara} analysis) and global-level (\textit{r\={\i}ti}, \textit{vakrokti}, \textit{rasa}, \textit{dhvani} and \textit{aucitya} analysis). As the names suggest, stanza-level analysis shows the analysis for each stanza and global analysis considers complete composition to provide the respective analysis. The stanza-level modules are shown in blue color and the rest in purple.  Our platform can provide the analysis in two possible ways. First, a user can choose to analyze any stanza through all the modules. Otherwise, the user can also opt for choosing any module and analyzing all the stanzas from the perspective of the chosen module. Our web template leverages the flex module to provide interactive features.

\section{Conclusion}
In conclusion, this article presents an innovative approach to bridge the gap between Sanskrit poetry and computational linguistics by proposing a framework that classifies Sanskrit poetry into levels of the characteristics of fine composition. The proposed framework takes a human-in-the-loop approach, combining the deterministic aspects delegated to machines and the deep semantics left to human experts. 
% This approach addresses the 5 computational challenges involved in the process, including subjectivity, complex language use, cultural context, lack of large labeled datasets, and multi-modality.
The article provides a deep analysis of \textit{\'Sik\d{s}\={a}\d{s}\d{t}aka}, a Sanskrit poem, from the perspective of 7 prominent \textit{\textit{k\={a}vya}\'s\={a}stra} schools to illustrate the proposed framework. The analysis establishes that \textit{\'Sik\d{s}\={a}\d{s}\d{t}aka} is indeed the \textit{uttama} \textit{k\={a}vya} composition. Moreover, the article provides compound, dependency, \textit{anvaya}, meter, \textit{rasa}, \textit{ala\.nk\=ara}, and \textit{r\={\i}ti} annotations for \textit{\'Sik\d{s}\={a}\d{s}\d{t}aka} and a web application to illustrate the poem's analysis and annotations.
The roadmap of the proposed framework opens new possibilities for the automatic analysis and classification of Sanskrit poetry while preserving the rich tradition of Sanskrit poetry.

\section*{Ethics Statement}
Our work proposes a roadmap framework to evaluate the aesthetic beauty of Sanskrit poetry, which we believe will be valuable to individuals interested in the analysis of Sanskrit poetry. We have taken great care to ensure that our analysis is conducted ethically and responsibly. We do not anticipate any adverse effects of our framework and analysis on any community. 
% Transparency and open access are fundamental principles that guide our work, and we believe that sharing our resources will benefit the wider Natural Language Processing (NLP) community.
% We are committed to continually considering the ethical implications of our framework's development and welcome feedback from the community on how to improve our ethical practices.

\section*{Acknowledgement}
We would like to express gratitude to several individuals who contributed to the successful completion of this research project. We are grateful to Chaitanya Lakkundi for his helpful discussions on grammar.
We would like to extend appreciation to Dr. Anupama Rayali, Dr. Pavankumar Salutari and Dr. Saurabh Dwivedi (Buddh P G College, Kushinagar, Uttar Pradesh) for their assistance in annotating \'Sik\d{s}\={a}\d{s}\d{t}aka. We are also thankful to Sugyan Kumar Mahanty for providing valuable inputs in their metrical analysis.
We thank Hrishikesh Terdalkar for helping us with the metrical analysis and web designing.
We acknowledge the support and assistance of Gauranga Darshan Prabhu, Dr. Amba Kulkarni (University of Hyderabad), Leela Govind Prabhu, Sriram Krishnan, Vedantasutra Prabhu and Venu Gopal Prabhu in connecting with different scholars of poetry school. We are also grateful to Shatavadhani R. Ganesh for his insightful discussions on clarifying our doubts.
We appreciate the anonymous reviewers for their insightful suggestions that helped in enhancing this work. The work of the first author is supported by the TCS Fellowship under Project TCS/EE/2011191P.

\bibliography{custom}
\bibliographystyle{acl_natbib}

\appendix

\section{\textit{\'Sik\d{s}\={a}\d{s}\d{t}aka} Composition}
\label{shikastakam_composition}
\begin{center}
  \noindent\textbf{Stanza 1}
  
ceto-darpa\d{n}a-m\={a}rjana\.m bhava-mah\={a}-d\={a}v\={a}gni-nirv\={a}pa\d{n}a\.m |

\'sreya\d{h}-kairava-candrik\={a}-vitara\d{n}a\.m vidy\={a}-vadh\={u}-j\={\i}vanam ||

\={a}nand\={a}mbudhi-vardhana\.m prati-pada\.m p\={u}r\d{n}\={a}m\d{r}t\={a}sv\={a}dana\.m |

sarv\={a}tma-snapana\.m para\.m vijayate \'sr\={\i}-k\d{r}\d{s}\d{n}a-sa\.nk\={\i}rtanam ||1|| \\  
\end{center}

\noindent\textbf{Synonyms:}
ceta\d{h} — of the heart; darpa\d{n}a — the mirror; m\={a}rjanam — cleansing; bhava — of material existence; mah\={a}-d\={a}va-agni — the blazing forest fire; nirv\={a}pa\d{n}am — extinguishing; \'sreya\d{h} — of good fortune; kairava — the white lotus; candrik\={a} — the moonshine; vitara\d{n}am — spreading; vidy\={a} — of all education; vadh\={u} — wife; j\={\i}vanam — the life; \={a}nanda — of bliss; ambudhi — the ocean; vardhanam — increasing; prati-padam — at every step; p\={u}r\d{n}a-am\d{r}ta — of the full nectar; \={a}sv\={a}danam — giving a taste; sarva — for everyone; \={a}tma-snapanam — bathing of the self; param — transcendental; vijayate — let there be victory; \'sr\={\i}-k\d{r}\d{s}\d{n}a-sa\.nk\={\i}rtanam — for the congregational chanting of the holy name of K\d{r}\d{s}\d{n}a.\\
\noindent\textbf{Translation:}
Let there be all victory for the chanting of the holy name of Lord K\d{r}\d{s}\d{n}a, which can cleanse the mirror of the heart and stop the miseries of the blazing fire of material existence. That chanting is the waxing moon that spreads the white lotus of good fortune for all living entities. It is the life and soul of all education. The chanting of the holy name of K\d{r}\d{s}\d{n}a expands the blissful ocean of transcendental life. It gives a cooling effect to everyone and enables one to taste full nectar at every step.\\

\begin{center}
  \noindent\textbf{Stanza 2}
  
n\={a}mn\={a}m ak\={a}ri bahudh\={a} nija-sarva-\'saktis |

tatr\={a}rpit\={a} niyamita\d{h} smara\d{n}e na k\={a}la\d{h} ||

et\={a}d\d{r}\'s\={\i} tava k\d{r}p\={a} bhagavan mam\={a}pi |

dur-daivam \={\i}d\d{r}\'sam ih\={a}jani n\={a}nur\={a}ga\d{h} ||2|| \\  
\end{center}

\noindent\textbf{Synonyms:}
n\={a}mn\={a}m — of the holy names of the Lord; ak\={a}ri — manifested; bahudh\={a} — various kinds; nija-sarva-\'sakti\d{h} — all kinds of personal potencies; tatra — in that; arpit\={a} — bestowed; niyamita\d{h} — restricted; smara\d{n}e — in remembering; na — not; k\={a}la\d{h} — consideration of time; et\={a}d\d{r}\'s\={\i} — so much; tava — Your; k\d{r}p\={a} — mercy; bhagavan — O Lord; mama — My; api — although; durdaivam — misfortune; \={\i}d\d{r}\'sam — such; iha — in this (the holy name); ajani — was born; na — not; anur\={a}ga\d{h} — attachment.\\
\noindent\textbf{Translation:}
My Lord, O Supreme Personality of Godhead, in Your holy name there is all good fortune for the living entity, and therefore You have many names, such as “K\d{r}\d{s}\d{n}a” and “Govinda,” by which You expand Yourself. You have invested all Your potencies in those names, and there are no hard and fast rules for remembering them. My dear Lord, although You bestow such mercy upon the fallen, conditioned souls by liberally teaching Your holy names, I am so unfortunate that I commit offenses while chanting the holy name, and therefore I do not achieve attachment for chanting.\\

\begin{center}
\noindent\textbf{Stanza 3}

t\d{r}\d{n}\={a}d api su-n\={\i}cena |

taror iva sahi\d{s}\d{n}un\={a} ||

am\={a}nin\={a} m\={a}na-dena |

k\={\i}rtan\={\i}ya\d{h} sad\={a} hari\d{h} ||3||
\end{center}

\noindent\textbf{Synonyms:}
t\d{r}\d{n}\={a}t api — than downtrodden grass; su-n\={\i}cena — being lower; taro\d{h} — than a tree; iva — like; sahi\d{s}\d{n}un\={a} — with tolerance; am\={a}nin\={a} — without being puffed up by false pride; m\={a}na-dena — giving respect to all; k\={\i}rtan\={\i}ya\d{h} — to be chanted; sad\={a} — always; hari\d{h} — the holy name of the Lord.\\
\noindent\textbf{Translation}
One who thinks himself lower than the grass, who is more tolerant than a tree, and who does not expect personal honor but is always prepared to give all respect to others can very easily always chant the holy name of the Lord.\\

\begin{center}
\noindent\textbf{Stanza 4}

na dhana\.m na jana\.m na sundar\={\i}\.m |

kavit\={a}\.m v\={a} jagad-\={\i}\'sa k\={a}maye || 

mama janmani-janmani \={\i}\'svare | 

bhavat\={a}d bhaktir a-haituk\={\i} tvayi ||4||\\ 
\end{center}
\noindent\textbf{Synonyms:}
na — not; dhanam — riches; na — not; janam — followers; na — not; sundar\={\i}m — a very beautiful woman; kavit\={a}m — fruitive activities described in flowery language; v\={a} — or; jagat-\={\i}\'sa — O Lord of the universe; k\={a}maye — I desire; mama — My; janmani — in birth; janmani — after birth; \={\i}\'svare — unto the Supreme Personality of Godhead; bhavat\={a}t — let there be; bhakti\d{h} — devotional service; ahaituk\={\i} — with no motives; tvayi — unto You.\\
\noindent\textbf{Translation:}
O Lord of the universe, I do not desire material wealth, materialistic followers, a beautiful wife or fruitive activities described in flowery language. All I want, life after life, is unmotivated devotional service to You.\\

\begin{center}
\noindent\textbf{Stanza 5}

ayi nanda-tanuja ki\.n-kara\.m |

patita\.m m\={a}\.m vi\d{s}ame bhava-ambudhau ||

k\d{r}pay\={a} tava p\={a}da-pa\.nkaja- |

sthita-dh\={u}l\={\i}-sad\d{r}\'sa\.m vicintaya||5||\\ 
\end{center}

\noindent\textbf{Synonyms:}
ayi — O My Lord; nanda-tanuja — the son of Nanda Mah\={a}r\={a}ja, K\d{r}\d{s}\d{n}a; ki\.nkaram — the servant; patitam — fallen; m\={a}m — Me; vi\d{s}ame — horrible; bhava-ambudhau — in the ocean of nescience; k\d{r}pay\={a} — by causeless mercy; tava — Your; p\={a}da-pa\.nkaja — lotus feet; sthita — situated at; dh\={u}l\={\i}-sad\d{r}\'sam — like a particle of dust; vicintaya — kindly consider.\\
\noindent\textbf{Translation:}
O My Lord, O K\d{r}\d{s}\d{n}a, son of Mah\={a}r\={a}ja Nanda, I am Your eternal servant, but because of My own fruitive acts I have fallen into this horrible ocean of nescience. Now please be causelessly merciful to Me. Consider Me a particle of dust at Your lotus feet.\\

\begin{center}
  \noindent\textbf{Stanza 6}
  
nayana\.m galad-a\'sru-dh\={a}ray\={a} |

vadana\.m gadgada-ruddhay\={a} gir\={a} ||

pulakair nicita\.m vapu\d{h} kad\={a} |

tava n\={a}ma-graha\d{n}e bhavi\d{s}yati||6||\\  
\end{center}

\noindent\textbf{Synonyms:}
nayanam — the eyes; galat-a\'sru-dh\={a}ray\={a} — by streams of tears running down; vadanam — mouth; gadgada — faltering; ruddhay\={a} — choked up; gir\={a} — with words; pulakai\d{h} — with erection of the hairs due to transcendental happiness; nicitam — covered; vapu\d{h} — the body; kad\={a} — when; tava — Your; n\={a}ma-graha\d{n}e — in chanting the name; bhavi\d{s}yati — will be.\\
\noindent\textbf{Translation:}
My dear Lord, when will My eyes be beautified by filling with tears that constantly glide down as I chant Your holy name? When will My voice falter and all the hairs on My body stand erect in transcendental happiness as I chant Your holy name?\\

\begin{center}
 \noindent\textbf{Stanza 7}
 
yug\={a}yita\.m nime\d{s}e\d{n}a |

cak\d{s}u\d{s}\={a} pr\={a}v\d{r}\d{s}\={a}yitam ||

\'s\={u}ny\={a}yita\.m jagat sarva\.m |

govinda-virahe\d{n}a me ||7||\\   
\end{center}

\noindent\textbf{Synonyms 7:}
yug\={a}yitam — appearing like a great millennium; nime\d{s}e\d{n}a — by a moment; cak\d{s}u\d{s}\={a} — from the eyes; pr\={a}v\d{r}\d{s}\={a}yitam — tears falling like torrents of rain; \'s\={u}ny\={a}yitam — appearing void; jagat — the world; sarvam — all; govinda — from Lord Govinda, K\d{r}\d{s}\d{n}a; virahe\d{n}a me — by My separation.\\
\noindent\textbf{Translation:}
My Lord Govinda, because of separation from You, I consider even a moment a great millennium. Tears flow from My eyes like torrents of rain, and I see the entire world as void.\\

\begin{center}
    \noindent\textbf{Stanza 8}
    
\={a}\'sli\d{s}ya v\={a} p\={a}da-rat\={a}\.m pina\d{s}\d{t}u m\={a}m |

a-dar\'san\={a}n marma-hat\={a}\.m karotu v\={a} ||

yath\={a} tath\={a} v\={a} vidadh\={a}tu lampa\d{t}o |

mat-pr\={a}\d{n}a-n\={a}thas tu sa eva n\={a}para\d{h} ||8||\\
\end{center}

\textbf{Synonyms:}
\={a}\'sli\d{s}ya — embracing with great pleasure; v\={a} — or; p\={a}da-rat\={a}m — who have fallen at the lotus feet; pina\d{s}\d{t}u — let Him trample; m\={a}m — Me; adar\'san\={a}t — by not being visible; marma-hat\={a}m — brokenhearted; karotu — let Him make; v\={a} — or; yath\={a} — as (He likes); tath\={a} — so; v\={a} — or; vidadh\={a}tu — let Him do; lampa\d{t}a\d{h} — a debauchee, who mixes with other women; mat-pr\={a}\d{n}a-n\={a}tha\d{h} — the Lord of My life; tu — but; sa\d{h} — He; eva — only; na apara\d{h} — not anyone else.\\
\textbf{Translation:}
Let K\d{r}\d{s}\d{n}a tightly embrace this maidservant who has fallen at His lotus feet, or let Him trample Me or break My heart by never being visible to Me. He is a debauchee, after all, and can do whatever He likes, but still He alone, and no one else, is the worshipable Lord of My heart.

\end{document}